\documentclass{article}
\usepackage{amssymb}

\usepackage[final]{corl_2025} 

\usepackage{multicol}
\usepackage{xcolor}
\usepackage{colortbl}
\usepackage{amsthm}
\usepackage{amsmath}
\usepackage{amssymb}
\usepackage{booktabs}    
\usepackage{array}
\usepackage{tabularx}
\usepackage{graphicx}
\usepackage{multirow}
\usepackage{amrl} 
\usepackage{subcaption}
\usepackage{wrapfig}  

\newcommand{\ourmethod}{ComposableNav}
\newcommand{\ourproblem}{Instruction-Following Navigation in Dynamic Environments}

\definecolor{prettyblue}{RGB}{0, 102, 204} 
\definecolor{purple}{RGB}{194, 113, 219}
\definecolor{lavander}{RGB}{232, 229, 243}
\usepackage[scaled=0.9]{DejaVuSansMono}

\usepackage{algorithm}
\usepackage{algorithmic}
\usepackage{listings}    
\newcommand{\hl}[1]{\textcolor{orange!80!black}{\textbf{#1}}}

\title{\ourmethod{}: \ourproblem{} via Composable Diffusion}

\author{
\textbf{Zichao Hu\textsuperscript{1}\thanks{Correspondence to: \texttt{zichao@utexas.edu}}, Chen Tang\textsuperscript{1}, Michael J.~Munje\textsuperscript{1}, Yifeng Zhu\textsuperscript{1}, Alex Liu\textsuperscript{1}, Shuijing Liu\textsuperscript{1}}\\
\textbf{Garrett Warnell\textsuperscript{1,2}, Peter Stone\textsuperscript{1,3}, Joydeep Biswas\textsuperscript{1}}\\[3pt]
\textsuperscript{1}Department of Computer Science, The University of Texas at Austin\\
\textsuperscript{2}Army Research Laboratory
\textsuperscript{3}Sony AI 
}

\begin{document}
\maketitle


\begin{abstract}
This paper considers the problem of enabling robots to navigate dynamic environments while following instructions. 
The challenge lies in the combinatorial nature of instruction specifications: each instruction can include multiple specifications, and the number of possible specification combinations grows exponentially as the robot’s skill set expands. For example, “overtake the pedestrian while staying on the right side of the road” consists of two specifications: \textit{``overtake the pedestrian"} and \textit{``walk on the right side of the road."}
To tackle this challenge, we propose \ourmethod{}, based on the intuition that following an instruction involves independently satisfying its constituent specifications, each corresponding to a distinct motion primitive. 
Using diffusion models, \ourmethod{} learns each primitive separately, then composes them in parallel at deployment time to satisfy novel combinations of specifications unseen in training. 
Additionally, to avoid the onerous need for demonstrations of individual motion primitives, we propose a two-stage training procedure: (1) supervised pre-training to learn a base diffusion model for dynamic navigation, and (2) reinforcement learning fine-tuning that molds the base model into different motion primitives.
Through simulation and real-world experiments, we show that \ourmethod{} enables robots to follow instructions by generating trajectories that satisfy diverse and unseen combinations of specifications, significantly outperforming both non-compositional VLM-based policies and costmap composing baselines. \footnote{Project page: \href{https://amrl.cs.utexas.edu/ComposableNav}{https://amrl.cs.utexas.edu/ComposableNav}}

\end{abstract}


\begin{figure*}[h]
    \centering
    \includegraphics[width=\linewidth]{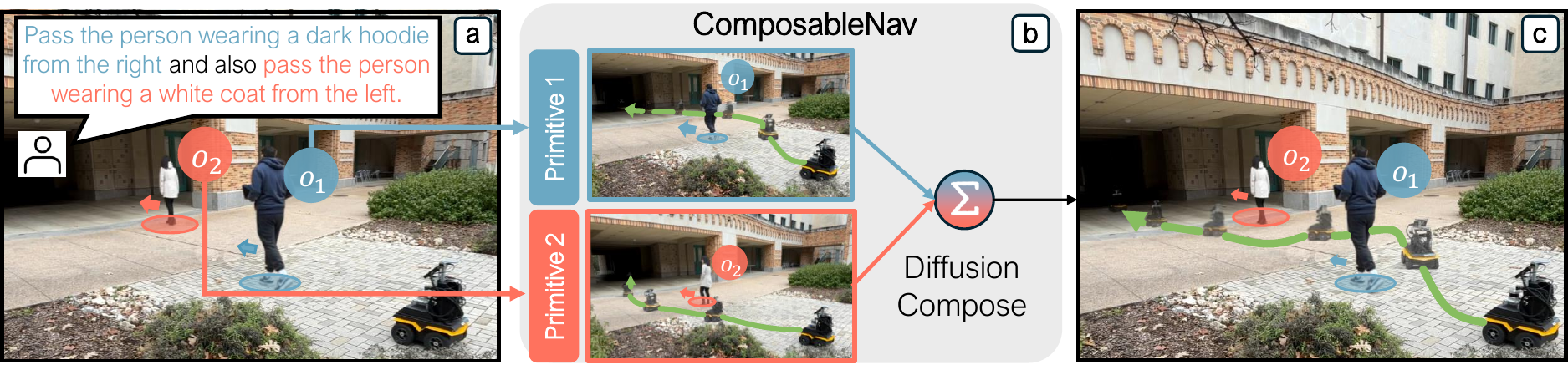}
    \caption{\textbf{
    \ourproblem{}.}
    Given an instruction that specifies how a robot should interact with entities in the scene (a), \ourmethod{} leverages the composability of diffusion models (b) to compose motion primitives to generate instruction-following trajectories (c). }
    \label{fig:first_figure}
\end{figure*}

\vspace{-1em}
\section{Introduction}
\vspace{-1em}


Developing robots that can effectively navigate by following instructions is an active research area aimed at enabling robots to operate within human-inhabited environments.
Existing work has predominantly tackled the instruction-following navigation problem in \emph{static} environments \cite{vln_survey1, vln_survey2, vln_survey3}, where instructions specify the navigation goals. In contrast, we focus on instruction-following navigation in \emph{dynamic} environments. Specifically, we consider the under-explored settings where the instructions describe specific robot interaction behaviors (\eg{} ``yield to a pedestrian") with respect to the other dynamic obstacles or agents. Addressing this problem requires developing methods capable of grounding high-level instructions into fine-grained, low-level actions that account for the dynamic behaviors of other agents. Solving this problem would allow end users (human or AI agents) to customize robotic behaviors beyond their default settings, in ways that align with user preferences and nuanced social interactions.

A crucial challenge present in instruction-following navigation is that a single instruction may contain multiple specifications for the robot to follow, e.g., \textit{``overtake the pedestrian while staying on the right side of the road”} consists of two specifications: \textit{``overtake the pedestrian"} and \textit{``walk on the right side of the road."}. Following an instruction amounts to simultaneously satisfying each of its constituent specifications. As the robot's capabilities and environment complexity increase, the space of possible combinations of such specifications grows exponentially. This combinatorial expansion makes popular learning-based methods, such as imitation learning~\cite{rt2} or reinforcement learning~\cite{rl_survey, flare}, impractical as they demand substantial data and computational resources. 

To address this challenge, we build our solution upon the idea of \textbf{\textit{composition}}. Rather than training a single model to handle an exponential number of possible combinations, we propose to train separate motion primitives for each individual category of specifications. At deployment time, we assume that an upstream module, such as a large language model, can decompose a natural-language instruction into a set of specifications online. The corresponding motion primitives can then be composed to generate a trajectory that follows the instruction. This approach significantly reduces complexity from exponential to linear: a relatively small set of motion primitives can support a combinatorially large space of instructions, enabling users to specify diverse robot behaviors required in real-world social navigation. Notably, in our setting, the relevant primitives are blended while composed, rather than stitched together sequentially~\cite{manipulation_sequential_compose1, manipulation_sequential_compose2}, since the robot's trajectory needs to satisfy all the specifications \emph{simultaneously}.


To this end, we present \textbf{\ourmethod{}}, a composable, diffusion-based motion planner that composes motion primitives based on the instruction specifications to generate instruction-following motion trajectories. The core intuition motivating our approach is that diffusion models~\cite{ddpm, diffusion-model} are highly effective at representing complex probability distributions, and that these models can be composed to form joint distributions~\cite{compositional_generation}. Leveraging this property, we can separately train diffusion models to learn motion primitives, each represented as a distribution over trajectories that satisfy a specific instruction specification. At deployment time, \ourmethod{} composes the relevant motion primitives based on the provided instruction specifications, constructs the corresponding joint distribution, and samples a trajectory that simultaneously satisfies all specified instructions.
Additionally, to avoid the onerous need for demonstrations of individual motion primitives, we introduce a two-stage training procedure consisting of supervised pre-training followed by reinforcement learning (RL) fine-tuning. Finally, to ensure real-time performance, we incorporate a model predictive controller (MPC)~\cite{mppi} and an online replanning strategy for low-latency action execution.

We demonstrate the effectiveness of \ourmethod{} through simulated and real-world experiments. With just six motion primitives (See \tabref{social_navigation_motion_primitive} in Appendix for the instruction list), we build a testbed with 24 instructions featuring various unseen specification combinations. Our results show that \ourmethod{} excels at following unseen instructions compared to baseline approaches. Our main contributions are summarized as follows.
(1) We introduce the use of \emph{composition} as a strategy for instruction-following navigation in dynamic environments, making the problem tractable under limited data and computational resources for training.
(2) We propose a diffusion-based learning method to model motion primitives as probability distributions, enabling their composition at deployment time.
(3) We develop a two-stage training procedure—combining supervised pre-training and reinforcement learning fine-tuning—that effectively learns motion primitives without the need for specialized demonstration datasets for each primitive.
\vspace{-0.8em}

\section{Related Work}
\vspace{-0.8em}
\textbf{Instruction Following Navigation.}
A key area in instruction-following navigation is vision-language navigation (VLN)\citep{vln_survey1, vln_survey2, vln_survey3}, which combines natural language understanding with visual perception to guide agents through 3D environments\citep{vln_method1, vln_method2, vln_method3, liu2024dragon, chang2023var, navid}. However, these methods assume static settings and overlook scenarios involving dynamic agents, where instructions specify interactions with moving agents. In contrast, social robot navigation focuses on enabling robots to operate in dynamic environments~\citep{idips, group_base_social_nav, layered_costmap_social_nav, DRL_VO, liu2020decentralized, sadrl, rethinking_social_nav, zheng2022hierarchical, height}, but lacks instruction conditioning. Recent work has shown promise in using vision-language models (VLMs) to address this gap. CoNVOI~\citep{convoi} and Social-VLM-Nav~\citep{social_vlm_nav} leverage VLMs' reasoning to interpret environment observations and suggest actions, but they face high inference latency and planning inconsistency. BehAV~\citep{behav} addresses some of these issues by generating cost maps from VLM outputs, yet struggles with sample inefficiency in geometric planning. Our work proposes a novel alternative using diffusion models~\cite{ddpm} to compose motion primitives, without relying on costly VLM inference.

\textbf{Diffusion For Robotics.}
Diffusion models have emerged as powerful tools for solving a variety of robotics tasks~\cite{3d_diffuser_actor, chi2023diffusionpolicy, MPD, black2024ddpo, seo2024presto, pbdiff}, with training typically performed using either supervised learning~\cite{chi2023diffusionpolicy, MPD, seo2024presto, 3d_diffuser_actor} or RL~\cite{black2024ddpo, dppo2024}. A distinctive advantage of diffusion models is their ability to guide the sampling process after training~\cite{reduce_reuse_recycle}. Janner et al.~\cite{janner2022diffuser} first relate this guided sampling mechanism to the control-as-inference framework~\cite{levine2018reinforcementlearningcontrolprobabilistic}, demonstrating that classifier guidance enables the generation of motion plans for previously unseen goal configurations. However, designing such classifiers can be challenging. To address this, Luo et al.~\cite{pbdiff} interpret diffusion models as energy-based models, training separate models and composing them at inference time to generalize to novel environments. Building on this idea of composition, our work differs in that we do not assume access to diverse motion primitive datasets for supervised training. Instead, we propose using reinforcement learning-based~\cite{black2024ddpo} to fine-tune diffusion models, and composing them to generate trajectories that satisfy unseen combinations of specifications from an instruction.

\section{Problem Formulation}
\label{sec:problem_formulation}
\vspace{-0.8em}

We consider the problem of instruction-following robot navigation in dynamic environments, where the objective is to generate a motion trajectory $\tau$ that follows a given instruction $I$, based on the robot's observation $O$ of the environment.
We represent the motion trajectory $\tau$ as a sequence of 2D waypoints at fixed-time intervals, which are then tracked by a model predictive controller to produce fine-grained actions in real time. The observation $O$ encodes the state of entities relevant to the instruction, such as the current and predicted positions of dynamic agents. Note that other representations are also possible, such as full SE(3) poses for $\tau$ or RGB images for $O$.

In this work, we assume an instruction $I$ can be decomposed into a set of independent specifications $I \rightarrow \langle \phi^{(1)}, \phi^{(2)}, \dots, \phi^{(k)} \rangle$. Each specification $\phi^{(i)}: \tau \times O \rightarrow \{0,1\}$ evaluates whether the trajectory meets the corresponding requirement, returning 1 if it does and 0 otherwise.
To determine whether a trajectory $\tau$ follows an instruction $I$, $\tau$ must satisfy all relevant specifications. Formally,
\begin{equation}
    \label{eq:problem_formulation}
    \tau \text{ follows } I \text{ iff }  \forall i \in [1,\cdots, k] , \phi^{(i)}(\tau, O) = 1.
\end{equation}
Solving this problem is challenging because the trajectory must simultaneously satisfy all specifications $\phi^{(i)}$, whose combinations can grow exponentially. In the following sections, we explain how leveraging  diffusion models enables us to compose motion primitives and generate trajectories that can follow instructions during robot deployment.

\section{Preliminaries}
\vspace{-0.8em}

We provide a brief overview of the two key techniques used in \ourmethod{}, conditional diffusion models~\cite{diffusion-model, ddpm, diffusion_beats_gan, ho2021classifierfree} and denoising diffusion policy optimization (DDPO)~\cite{black2024ddpo}.

\subsection{Conditional Diffusion Models}
\vspace{-0.8em}

In this work, we consider conditional diffusion probabilistic models~\cite{ddpm, diffusion-model, ho2021classifierfree}, which belong to a family of generative models trained to represent a conditional distribution $p(x\mid c)$, where $c$ is the corresponding context. 
These models are trained to reverse a forward diffusion process $q(x_t\mid x_{t-1})$ that gradually adds Gaussian noise to the data $x_0 \sim p(x|c)$. To learn this reverse process, the model is trained to predict the noise $\epsilon$ at each step $t$ using a denoising network, $f_{\theta}(x_t, t, c) \approx \epsilon$, where $x_t$ is the noisy data at step $t$.
 The network is optimized using a training objective that penalizes the mean squared error between the predicted and actual noise value at step $t$: 
\begin{equation}\label{eq:diffusion_training_objective}
    \mathcal{L}_\text{MSE}(\theta) = \mathbb{E}_{x_0, \epsilon, t, c} \left[ \|\epsilon - f_{\theta}(x_t, t, c) \|^2 \right].
\end{equation}
This objective is derived from maximizing a variational lower bound on the data log-likelihood~\cite{ddpm}.

At inference time, the model generates a data sample by starting from Gaussian noise $x_T \sim \mathcal{N}(0, \mathbf{I})$ and progressively denoising it using the learned denoising network for $T$ steps. 
The reverse process at each timestep $t$ follows a Gaussian distribution with a time-dependent covariance matrix $\sigma_t^2\mathbf{I}$, where $\sigma_t^2$ is treated as a hyperparameter: 
\begin{equation} 
\label{eq:denoise2} 
p_\theta(x_{t-1} \mid x_t, c) = \mathcal{N}(x_t - f_{\theta}(x_t, t, c), \sigma_t^2\mathbf{I}), 
\end{equation} 
This iterative process continues until a final sample $x_0$ is obtained, which approximates the true conditional distribution $p(x\mid c)$.

\subsection{Denoising Diffusion Policy Optimization (DDPO)}
\label{sec:ddpo_prelim}
\vspace{-0.8em}

\ourmethod{} follows the denoising diffusion policy optimization technique (DDPO) proposed by Black et al.~\citep {black2024ddpo} to use reinforcement learning (RL) to fine-tune diffusion models to generate the motion primitives corresponding to the instruction specifications. 
DDPO models the multi-step denoising process as a multi-step Markovian Decision Process (MDP), defined as a tuple $\mathcal{M} = \langle \mathcal{S}, \mathcal{A}, \rho_0, \mathcal{P}, R\rangle$, where $\mathcal{S}$ is the state space, $\mathcal{A}$ is the action space, $\rho_0$ is the distribution of initial states, $\mathcal{P}$ is the transition kernel, and $R$ is the reward function. We denote the timestep of this multi-step MDP as $i$. The denoising process is mapped into this MDP as follows: 
\begin{equation}
\begin{array}{lll}
s_{i} \triangleq \langle x_t, t, c \rangle \in \mathcal{S} &
\pi(a_{i} \mid s_{i}) \triangleq p_\theta(x_{t-1} \mid x_t,c) &
P(s_{i+1} \mid s_{i}, a_{i}) \triangleq \langle \delta_{x_{t-1}}, \delta_{t-1}, \delta_c \rangle \\[1em]
a_{i} \triangleq x_{t-1} \in \mathcal{A} &
\rho_0(s_0) \triangleq \langle \mathcal{N}(0,\mathbf{I}), \delta_T, p(c) \rangle &
R(s_{i}, a_{i}) \triangleq 
\begin{cases} 
    r(x_0, c) & \text{if } t = 0, \\
    0 & \text{otherwise},
\end{cases}
\end{array}
\end{equation}
where $\delta_y$ denotes the Dirac distribution with nonzero density only at $y$. 

The key insight behind this technique is that the reverse process in a diffusion model is a Markovian process, where each denoising step $p_\theta(x_{t-1} \mid x_t, c)$ is modeled as a Gaussian distribution (see \eqref{denoise2}). By interpreting each denoising step as the policy $\pi(a_{i} \mid s_{i})$ in an MDP, the policy itself becomes Gaussian, which allows for the exact evaluation of log-likelihoods and their gradients with respect to the diffusion model parameters.
As a result, this formulation enables the use of policy gradient methods, such as PPO~\cite{ppo}, to optimize the diffusion model’s denoising network. 

The DDPO algorithm alternates between (1) collecting denoising trajectories $\langle x_T, x_{T-1}, \ldots, x_0 \rangle$ via sampling and (2) updating the model parameters using gradient descent. Finally, the policy gradient objective used in DDPO can be expressed as:
\begin{equation} 
\mathcal{L} = \mathbb{E} \left[ \sum_{t=1}^{T} \frac{p_\theta(x_{t-1} \mid x_t, c)} {p_{\theta_{\text{old}}}(x_{t-1} \mid x_t, c)} \nabla_\theta \log p_\theta(x_{t-1} \mid x_t, c) r(x_0, c) \right], 
\end{equation} 
where the expectation is taken over trajectories generated using the previous model parameters $\theta_{\text{old}}$.

\section{ComposableNav}
\vspace{-0.8em}

\begin{figure*}[h]
    \centering
    \includegraphics[width=\linewidth]{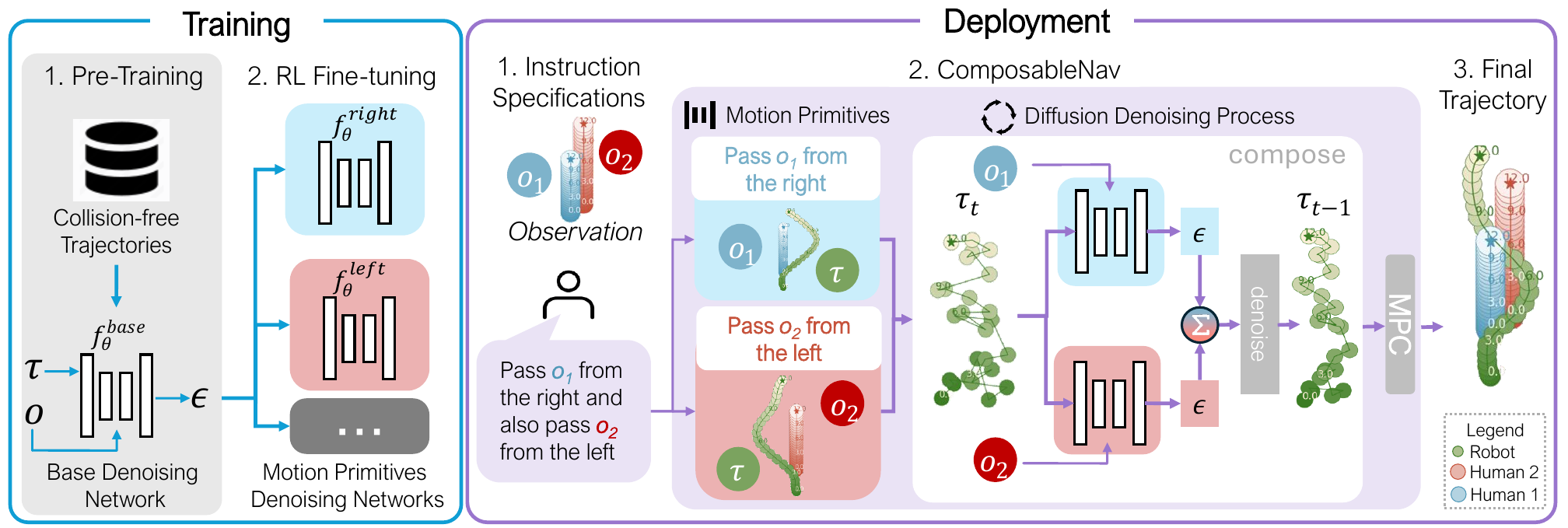}
    \caption{\textbf{\ourmethod{} Overview.} \ourmethod{} learns each motion primitive through a two-stage procedure: supervised pre-training (step 1) followed by RL fine-tuning (step 2). At deployment, it maps instruction specifications to the corresponding learned primitives and composes them by summing the predicted noise from each primitive’s denoising network. The final trajectory is generated through the diffusion denoising process.
    }
    \label{fig:main_figure}
    \vspace{-2em}
\end{figure*}

In this section, we present \ourmethod{}, a diffusion-based planner for instruction-following navigation. As shown in \figref{main_figure}, \ourmethod{} first learns motion primitives via a two-stage training procedure (see \secref{learn_primitive}). At deployment, given instruction specifications, it selects relevant primitives and composes them by summing the predicted noise from each diffusion model during the denoising process (\secref{compose_primitive}). Finally, for real-time control, \ourmethod{} is paired with an MPC (see Appendix~\ref{sec:real_time_deploy}). Please refer to our codebase for detailed implementation. \footnote{Code is released at https://github.com/ut-amrl/ComposableNav.}




\subsection{Learning Motion Primitives without Primitive-Specific Demonstration Data}
\label{sec:learn_primitive}
\vspace{-0.8em}

Diffusion models for robotics are often trained in a supervised manner using large-scale demonstration datasets~\cite{pbdiff,3d_diffusion_policy, chi2023diffusionpolicy}. However, collecting such datasets for different robot motion primitives can be labor-intensive. To address this challenge, we make two key design choices. 
First, \emph{dedicated datasets for each primitive behavior are costly to build, whereas general-purpose navigation datasets are far easier to obtain}. These general-purpose datasets only need to provide diverse, collision-free, goal-reaching trajectories in dynamic environments, which can be obtained from existing real-world datasets~\cite{scand,coda} or generated in simulation~\cite{MPD}.
Such data allows the pre-training of a base diffusion planner to generate diverse and feasible trajectories across various environments. 
Second, \emph{verifying a solution is often easier than generating one}. In our setting, evaluating whether a trajectory satisfies an instruction (\eg{} with rule-based heuristics or vision-language models (VLMs)) is typically more straightforward than directly generating such a trajectory. Hence, for each primitive, we design a primitive-specific reward function that asserts instruction compliance and use RL to fine-tune the pre-trained base model~\cite{black2024ddpo}. Building on the two design choices above, we propose a two-stage framework: supervised pretraining followed by RL fine-tuning.

\textbf{Supervised Pre-training.}
To pre-train a base diffusion model, we generate diverse trajectory data in simulation for simplicity and scalability. Following prior works~\cite{pbdiff, MPD}, we randomly synthesize environments with varying entities (\eg{} dynamic agents or terrain regions) and goal locations. We then use a geometric planner to generate a diverse set of collision-free, goal-reaching, and smooth trajectories. To account for dynamic environments, these trajectories must be \emph{time-dependent}, so we employ a spatio-temporal Hybrid A* planner. In addition, we also want to capture the distribution of diverse feasible trajectories within the same environment (\eg{} both detouring left or right around an obstacle in front are feasible trajectories). Hence, we use a Rapidly-Exploring Random Tree planner to randomly generate candidate trajectories and then select waypoints along the trajectories as subgoals for Hybrid A* to track. Additionally, we also vary the hyperparameters for the planners (\eg{} velocity cost) to further enhance trajectory diversity.

Once a diverse set of time-dependent trajectories is generated, we pre-train a base diffusion model via supervised learning, using the objective in \eqref{diffusion_training_objective}. The model learns a conditional denoising network $f^{\text{(base)}}_\theta (\tau_t, t, O)$, which predicts the noise $\epsilon$ to denoise the trajectory $\tau_t$ at step $t$, conditioned on environment observations $O$. We adopt an object-centric representation for the observations, encoding each observation separately and then using a transformer encoder to attend over these embeddings to produce a global context feature. To handle varying trajectory lengths, we pad shorter trajectories with the final goal position to ensure uniform length during training.

\textbf{RL Fine-tuning.} We then fine-tune the base model \emph{separately} for each motion primitive using RL, following the DDPO approach described in \secref{ddpo_prelim}. 
For each primitive, we randomly generate simulation environments containing only the entities relevant to the corresponding instruction specification. The diffusion model then generates trajectories for these environments, which are evaluated using a reward function based on how well they align with the instruction.
While the reward function can take various forms, we adopt a simple rule-based heuristic approach, as the primitives considered in our experiments are straightforward to evaluate (example shown in Appendix~\ref{sec:motion_primitives}). The resulting trajectories and rewards are stored in a replay buffer, and the model is updated using PPO~\cite{ppo}. Finally, after fine-tuning, we obtain multiple diffusion models $f^{\phi^{(i)}}_\theta (\tau_t, t, O)$, each representing a motion primitive associated with a specification $\phi^{(i)}$.

\subsection{Generating Instruction-Following Trajectories via Composing Motion Primitives}
\label{sec:compose_primitive}
\vspace{-0.8em}


\ourmethod{} models the instruction-following motion trajectories $
\tau$ as a conditional distribution $p(\tau| \phi^{(1)}, o^{(1)}, \cdots, \phi^{(k)}, o^{(k)})$, where each $\phi^{(i)}$ is a specification extracted from the instruction $I$, \textit{i.e.,}  $I \rightarrow \langle \phi^{(1)}, \cdots, \phi^{(k)}\rangle$, and each $o^{(i)}$ is the environment observation corresponding to $\phi^{(i)}$. 
We assume both the specifications and the environment observations can be extracted using off-the-shelf large language models and vision foundation models.
Given the conditional independence assumption for each specification $\phi^{(i)}$ discussed in \secref{problem_formulation}, the conditional distribution can be factorized as follows (derivation shown in \eqref{derivation}):
\begin{equation}
    \label{eq:condition_factor}
    p(\tau | \phi^{(1)}, o^{(1)}, \cdots, \phi^{(k)}, o^{(k)}) \propto p(\tau) \prod_{i=1}^{k} \frac{p(\tau \mid \phi^{(i)}, o^{(i)})}{p(\tau)}.
\end{equation}
Here, each conditional trajectory distribution $p(\tau \mid \phi^{(i)}, o^{(i)})$ corresponds to a motion primitive represented by a diffusion model with denoising network $f_\theta^{\phi^{(i)}}(\tau_t, t, o^{(i)})$. In contrast, the marginal trajectory distribution $p(\tau)$ is an unconditioned motion primitive, obtained by replacing the observation $o^{(i)}$ with a null input $\varnothing$, \textit{i.e.,} $f_\theta^{\phi^{(i)}}(\tau_t, t, \varnothing)$, following the classifier-free guidance approach~\cite{ho2021classifierfree}.

Following prior work~\cite{compositional_generation, pbdiff}, we compose motion primitives by summing the predicted noise from denoising networks, with user-defined weights $w_i$ controlling the guidance strength for the $i$th primitive ($w_i$ is set to be 1 for all primitives in this work). The composed noise is:
\begin{equation}
    \hat{\epsilon} = \frac{1}{k}\sum_{i=1}^{k} f^{\phi^{(i)}}_\theta(\tau_t, t, \varnothing) + \sum_{i=1}^k w_i(f^{\phi^{(i)}}_\theta(\tau_t, t, o^{(i)}) - f^{\phi^{(i)}}_\theta(\tau_t, t, \varnothing)).
\end{equation}
Here, with separate diffusion models for each primitive, $p(\tau)$ is defined as the average of the unconditioned outputs $f^{\phi^{(i)}}_\theta(\tau_t, t, \varnothing)$ across all considered primitives.

Finally, \ourmethod{} generates trajectories by iteratively applying the reverse diffusion process, starting from a noisy trajectory $\tau_T \sim \mathcal{N}(0, \mathbf{I})$ and denoising via $p_\text{compose}(\tau_{t-1} \mid \tau_t, \phi^{(1)}, o^{(1)}, \dots, \phi^{(k)}, o^{(k)}) = \mathcal{N}(\tau_t - \hat{\epsilon}, \sigma_t^2\mathbf{I})$. After $T$ steps, the process yields a trajectory $\tau_0$, drawn from a distribution concentrated on trajectories that satisfy all specifications of the given instruction (See Appendix~\ref{sec:compose_diffusion_section} for an intuitive explanation of diffusion composition, based on the score-based interpretation~\cite{song2021scorebased}).



\section{Experiments and Results}
\vspace{-0.8em}

We evaluate \ourmethod{} in simulation and the real world to address the following questions: (1) Can \ourmethod{} learn individual motion primitives that satisfy each instruction specification without relying on demonstration data? (2) To what extent can \ourmethod{} compose motion primitives to generate trajectories that satisfy unseen combinations of specifications, in comparison to baseline approaches? (3) Can \ourmethod{} operate in real-time when deployed on a real-world robot and enable the robot to follow instructions in dynamic environments involving pedestrian interactions? In our experiments, we consider six navigation motion primitives, as shown in \figref{primitives}~and Appendix~\ref{sec:motion_primitives}, with training details provided in Appendix~\ref{sec:training_detail}.


\begin{figure}[h]
    \centering

    \begin{subfigure}[t]{\textwidth}
        \centering
        \includegraphics[width=\textwidth]{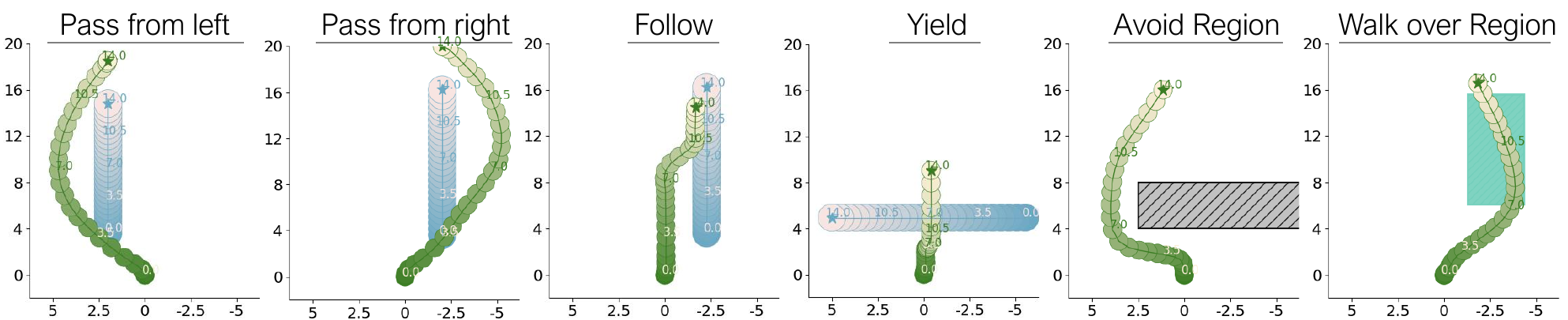}
        \caption{\textbf{Motion Primitives Learned in \ourmethod{}.} }
        \label{fig:primitives}
    \end{subfigure}

    \begin{subfigure}[t]{\textwidth}
        \centering
        \includegraphics[width=\textwidth]{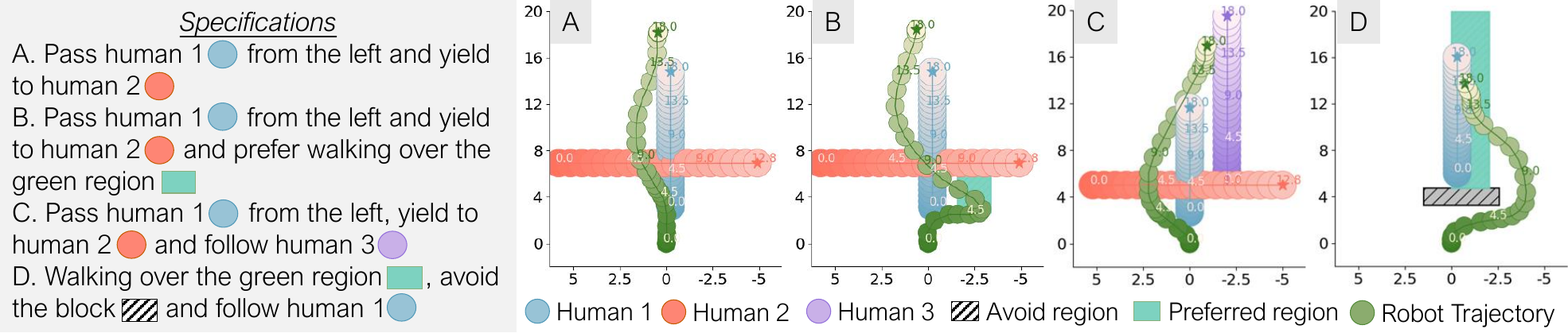}
        \caption{\textbf{Composed Robot Navigation Trajectories.} These examples show trajectories generated by combining learned motion primitives to follow complex instructions across diverse scenarios.}
        \label{fig:composed_experiment}
    \end{subfigure}

    \caption{Qualitative Simulation Results.}
    \label{fig:simulation_experiment}
    \vspace{-2em}
\end{figure}

\subsection{Simulation Experiments}
\vspace{-0.8em}
\textbf{Environment Setup.}
Using six motion primitives, we built a testbed with 24 instructions featuring various unseen specification combinations (see Appendix~\ref{sec:full_simulation_quantitative}). Instructions are grouped by complexity, based on the number of specifications—ranging from two to four—with each category comprising eight instructions. For each instruction, we randomly generate 20 test environments. \figref{composed_experiment}~illustrates sample environments and corresponding robot behaviors, where dynamic humans are modeled as spheres and regions as rectangles.

\textbf{Baselines.}
We compare \ourmethod{} with three VLM-based baselines (see Appendix~\ref{sec:baseline_details} for more details): (1) VLM-Social-Nav~\citep{social_vlm_nav}, which uses a VLM to select actions from predefined behaviors and translate them into a social cost function; (2) CoNVOI~\citep{convoi}, which leverages a VLM to choose the robot's next waypoint from an annotated image; and (3) BehAV~\citep{behav}, which uses a VLM to generate instruction-grounded segmentation maps converted into cost maps for motion planning. Like \ourmethod{}, these methods pair their approach with a local motion planner~\cite{dwa}.

\begin{wrapfigure}{r}{0.5\textwidth}
    \centering
    \includegraphics[width=\linewidth]{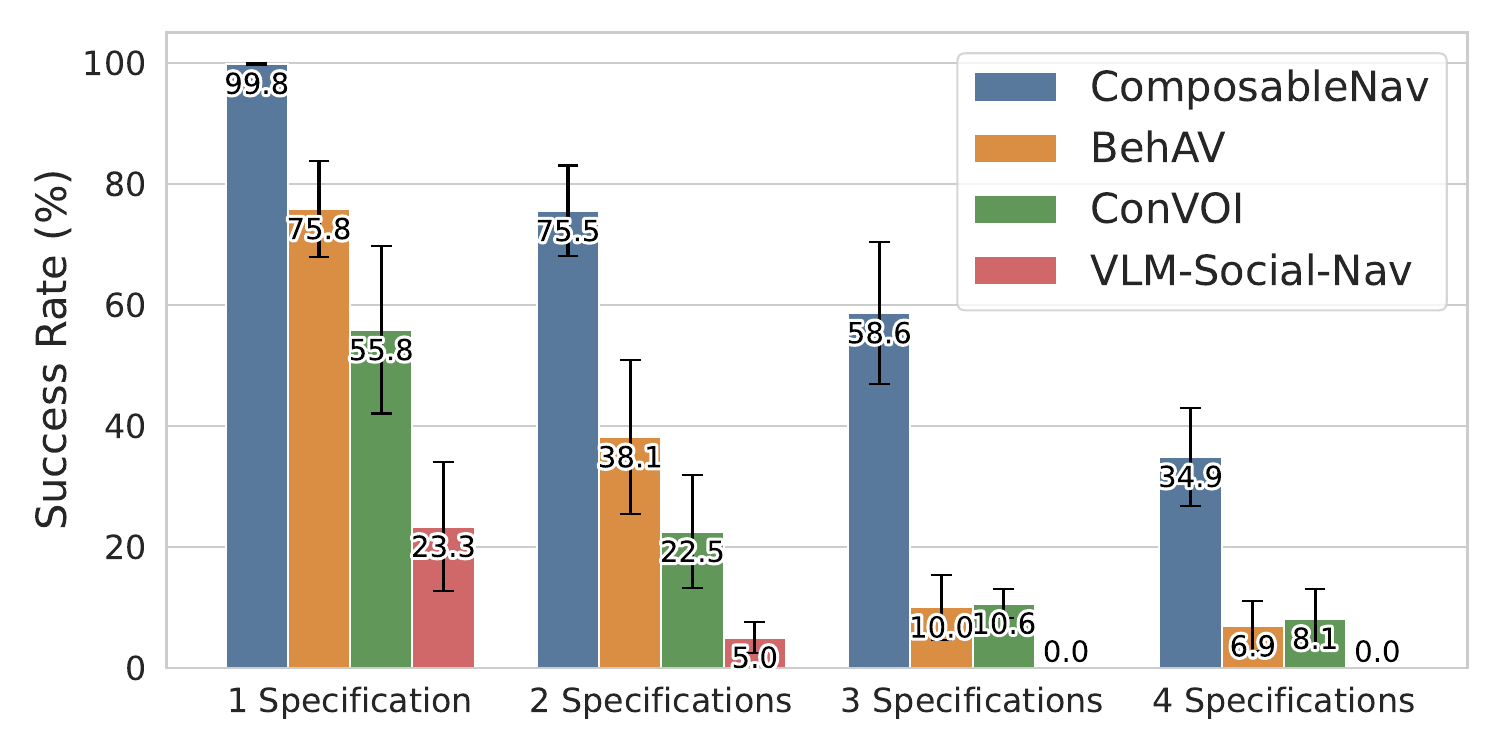}
        \caption{\textbf{Simulation Evaluation Results.} Bar plots show mean success rates with standard error bars. As instruction complexity increases, \ourmethod{} consistently outperforms baselines.}
    \label{fig:simulation_quantitative}
\end{wrapfigure}

\textbf{Metrics.}
Success is defined using three criteria—Instruction Alignment (IA), Collision-Free (CF), and Goal Reaching (GR)—with a trajectory considered successful (SR=1) only if all are satisfied. IA is assessed via a rule-based function to check if the trajectory follows the given instructions (See Appendix~\ref{sec:motion_primitives}).



\textbf{Results \& Analysis.} 
We first assess whether our two-stage training procedure enables diffusion models to learn individual motion primitives without direct demonstration data.
As illustrated in \figref{primitives}, the robot can generate navigation behaviors tailored to different primitives and environments. Also, the fine-tuned models execute motion primitives with near-perfect accuracy, whereas pre-trained models perform significantly worse (see \tabref{pretrain_finetune} in Appendix).

\figref{simulation_quantitative}~presents the quantitative results, with further analysis in Appendix~\ref{sec:full_simulation_quantitative}. \ourmethod{} consistently outperforms all baselines across all levels of instruction complexity. While the baseline methods exhibit moderate performance with a single specification, their success rates degrade rapidly as the number of specifications increases. In contrast, \ourmethod{} maintains fairly high success rates even under complex multi-specification instructions, achieving 58.6\% with three and 34.9\% with four specifications, where all baselines fall below 11\%. This demonstrates the robustness of ComposableNav in following complex unseen combinations of instruction specifications. 


Finally, \figref{composed_experiment}~further illustrates how \ourmethod{} composes motion primitives to generate diverse trajectories in response to previously unseen combinations of instruction specifications. For instance, to follow instruction \textit{A}, the trajectory must first slow down to yield to Person 2, then accelerate to overtake Person 1. \textit{B} differs from \textit{A} by introducing an additional specification to walk over the green region, prompting the robot to take a detour to the right. Similarly, \textit{C} and \textit{D} elicit complex navigation behaviors, such as following a human or avoiding a region in front.

\vspace{-0.7em}
\subsection{Deployment on a Real Robot}
\vspace{-0.7em}
\begin{figure}[h]
    \centering
    \includegraphics[width=\linewidth]{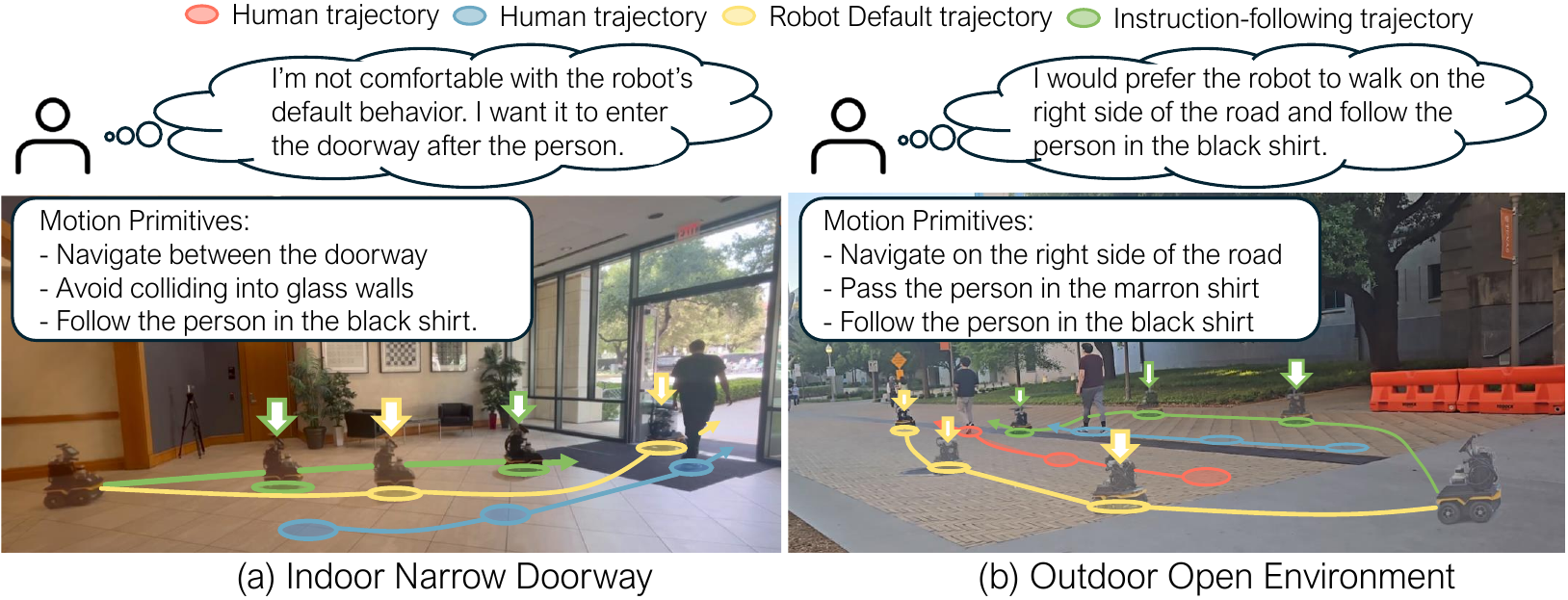}
    \caption{\textbf{Real world Experiments}. \ourmethod{} allows for customizing the robot’s behavior. }
    \label{fig:real-world-exp}
    \vspace{-1em}
\end{figure}

\begin{wrapfigure}{r}{0.35\textwidth} 
    \centering
    \includegraphics[width=\linewidth]{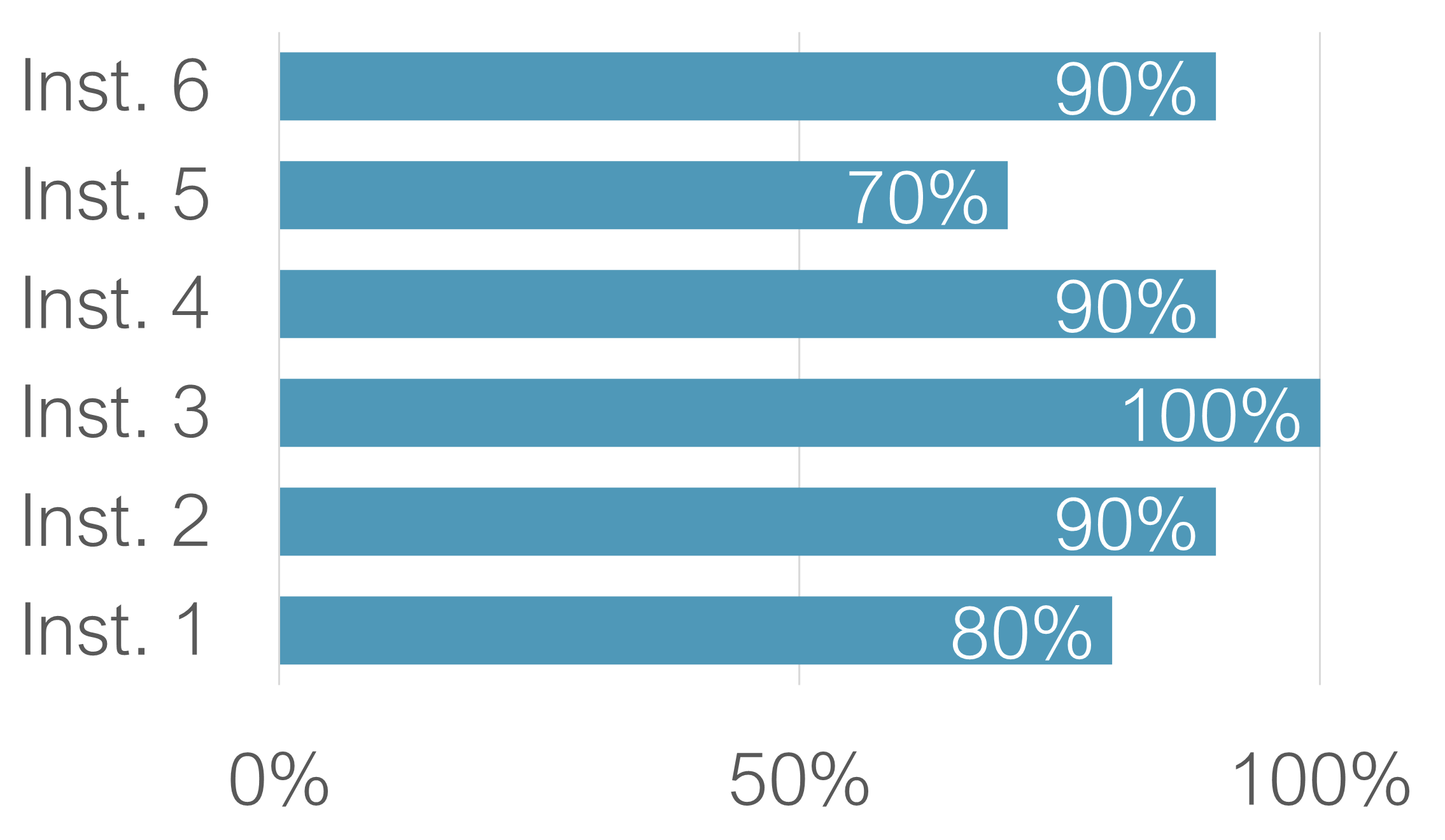}
    \captionof{figure}{Real-World Deployment Success Rate}
    \label{fig:robot_quantitative}
\end{wrapfigure}

We deployed \ourmethod{} on a Clearpath Jackal robot using only its onboard compute in real-world environments (see Appendix~\ref{sec:robot_setup}). We tested six instructions in two common scenarios: (1) navigating a narrow doorway and (2) walking in an open outdoor space. To ensure safety, all instructions were first validated in simulation. Each instruction was tested in 10 repeated trials to evaluate real-world performance. Success rates, shown in \figref{robot_quantitative}, were consistently high, indicating that \ourmethod{} can be effectively deployed in real-world settings (additional results detailed in Appendix~\ref{sec:additional_robot_deployment}).

\figref{real-world-exp}~also illustrates how \ourmethod{} enables users to customize robot behavior beyond the default navigation behavior. For example, the default geometric stack (yellow trajectory) prioritizes fast, collision-free navigation but may pass uncomfortably close to humans. In contrast, \ourmethod{} allows users to specify preferences, such as following a person through a doorway instead of overtaking them (green trajectory), supporting more human preference-aligned behavior.
Finally, we measured system latency, which depends on the number of composed motion primitives.  
In the most complex case involving four primitives, initial trajectory generation averaged 0.4s, whereas trajectory replanning required only 0.06s (see additional results in Appendix~\ref{sec:latency}).
\section{Conclusion}
\vspace{-0.8em}
This paper presents \ourmethod{}, a composable diffusion-based motion planner that composes navigation motion primitives to generate trajectories satisfying diverse, previously unseen combinations of instruction specifications. To address the lack of primitive-specific demonstrations, \ourmethod{} employs a two-stage training pipeline: (1) supervised pre-training to learn a base diffusion model for dynamic navigation, and (2) reinforcement-learning fine-tuning that specializes this base model for each motion primitive. Our simulation and real-world experiments show that the design of \ourmethod{} enables it to learn and compose motion primitives effectively, and it can be deployed on real-world robots to operate in real time.

\section{Limitations and Future Work}
This work has several limitations that future research could address.

First, we only considered six commonly used navigation primitives when composing novel behaviors. These primitives are relatively simple and can be described with straightforward rule-based reward functions (see~\secref{motion_primitives}). However, manually designing such reward functions does not scale well as the number of primitives grows. Importantly, our method is not tied to a particular way of crafting rewards. A promising and direct direction for improving scalability is to leverage vision-language models (VLMs) as verifiers—shown effective in DDPO~\cite{black2024ddpo}—to automatically learn diverse and complex behaviors without relying on handcrafted rewards.

Second, we assume that tasks such as parsing instructions into specifications and detecting relevant observations can be handled by existing methods, such as off-the-shelf LLMs and VLMs. Since our focus is on composable planning, we abstract these components away in our experiments. Future work may stack high-level VLM-based modules, as in prior studies~\cite{behav, huang2023voxposer}, on top of \ourmethod{} to close the loop. Furthermore, a high-level task planner~\cite{liu2023llmp, codebotler} could be introduced to further support long-horizon instruction-following navigation. 

Third, although \ourmethod{} significantly outperforms baseline methods, it still shows a notable decline in success rate as the number of instruction specifications increases. This limitation may stem from the underlying composition strategy: following Liu et al.~\cite{compositional_generation}, our method composes diffusion models by summing the predicted noise from individual denoising networks. As Du et al.~\cite{reduce_reuse_recycle} highlight, however, this approach can lead to suboptimal results. Future work may explore more advanced sampling techniques for diffusion models—such as Hamiltonian Monte Carlo~\cite{hmc1, hmc2}—to improve composition performance under higher instruction complexity.






\acknowledgments{
This work has taken place in the Autonomous Mobile Robotics Laboratory (AMRL) and the Learning Agents Research
Group (LARG) at UT Austin. AMRL research is supported in part by NSF (CAREER-2046955, OIA-2219236, DGE-2125858, CCF-2319471), ARO (W911NF-23-2-0004), Amazon, and JP Morgan. Any opinions, findings, and conclusions expressed in this material are those of the authors and do not necessarily reflect the views of the sponsors. LARG research is supported in part by NSF
(FAIN-2019844, NRT-2125858), ONR (N00014-24-1-2550), ARO (W911NF-17-2-0181, W911NF-23-2-0004, W911NF-25-1-0065), DARPA
(Cooperative Agreement HR00112520004 on Ad Hoc Teamwork) Lockheed Martin, and UT Austin's Good Systems grand challenge.  Peter Stone serves as the Chief Scientist of Sony AI and receives financial compensation for that role.  The terms of this arrangement have been reviewed and approved by the University of Texas at Austin in accordance with its policy on objectivity in research.}

\bibliography{references}  

\clearpage
\appendix


\section{Derivation for \eqref{condition_factor}}
\begin{align}
    \label{eq:derivation}
    &p(\tau | \phi^{(1)}, o^{(1)}, \cdots, \phi^{(k)}, o^{(k)}) \\
    &= \frac{p(\tau, \phi^{(1)}, o^{(1)}, \cdots, \phi^{(k)}, o^{(k)})}{p(\phi^{(1)}, o^{(1)}, \cdots, \phi^{(k)}, o^{(k)})} \tag*{\(\triangleleft\) Bayes' Rule}\\
    &\propto p(\tau, \phi^{(1)}, o^{(1)}, \cdots, \phi^{(k)}, o^{(k)}) \nonumber \\
    &= p(\tau) p(\phi^{(1)}, o^{(1)}, \cdots, \phi^{(k)}, o^{(k)} \mid \tau)   \tag*{\(\triangleleft\) Bayes' Rule}\\
    &=  p(\tau) \prod_{i=1}^{k} p(\phi^{(i)}, o^{(i)} \mid \tau)  \tag*{\(\triangleleft\) Conditional Independence}\\
    &\propto p(\tau) \prod_{i=1}^{k} \frac{p(\tau \mid \phi^{(i)}, o^{(i)})}{p(\tau)} \tag*{\(\triangleleft\) Bayes' Rule}
\end{align}

\section{Composing Diffusion Models via Score Function Interpretation}
\label{sec:compose_diffusion_section}
Diffusion models belong to the family of score-based generative models. These models learn to estimate the score function~\cite{song2021scorebased, compositional_generation}, which is defined as the gradient of the log-probability density with respect to the input, i.e., $\nabla_x \log p(x)$. \emph{Intuitively, the score function indicates the direction in which a data point should be moved to increase its likelihood under the data distribution.}

In the case of diffusion models, the denoising network $f_\theta(x_t, t)$ can be interpreted as being proportional to the score function. The denoising process can thus be viewed as iteratively moving the noisy sample $x_t$ in the direction predicted by the model, gradually transforming it into a high-probability sample from the data distribution. 

Given this interpretation, composing multiple diffusion models corresponds to computing the sum of their score functions, $\sum_{i=1}^k f^{(i)}_\theta(x_t, t)$, where each $f^{(i)}_\theta$ represents the score from the $i$-th diffusion model being composed and $\hat{\epsilon}$ is the composed score. The generative process for composing these models becomes~\cite{compositional_generation}:
\begin{equation}
    \label{eq:compose_diffusion_equation}
    p_\theta(x_{t-1} \mid x_t) = \mathcal{N}(x_t - \sum_{i=1}^k f^{(i)}_\theta(x_t, t), \sigma_t^2\mathbf{I})
\end{equation}
This process can be understood as guiding the sample toward regions that are simultaneously high-probability under all the models being composed. Hence, the data can be seen as sampling from the joint distribution defined by all the composed diffusion models.

\section{Motion Primitives}
\label{sec:motion_primitives}
In this work, we consider six commonly used navigation motion primitives, as listed in \tabref{social_navigation_motion_primitive}. In the following subsections, we present the pseudocode that outlines how we define the specification $\phi^{(i)}$ for each motion primitive using heuristic rule-based functions.

\begin{table}[H]
\centering
\caption{Instruction Specifications for Navigation Motion Primitives}
\label{tab:social_navigation_motion_primitive}
\footnotesize 
\begin{tabularx}{\linewidth}{>{\hsize=0.3\hsize\arraybackslash}X|>{\hsize=0.7\hsize\arraybackslash}X}
\toprule
\textbf{Motion Primitive (MP)} & \textbf{Instruction Specification} \\
\midrule
Pass a person from the left (L) & The robot should pass the person from the left side. \\
\midrule
Pass a person from the right (R) & The robot should pass the person from the right side. \\
\midrule
Follow behind a person (F) & The robot should stay in a specific region behind the person relative to the person's position. \\
\midrule
Yield to a person (Y) & The robot should not cross the region in front of the person. \\
\midrule
Walk through a region (W) & The robot's trajectory should overlap with the specified region. \\
\midrule
Avoid walking through a specified region (A) & The robot's trajectory should not overlap with the specified region, which may be defined either by a terrain feature or by a person's location. \\
\bottomrule
\end{tabularx}
\end{table}

\subsection{Pass a person from the left}
\begin{algorithm}[H]
\caption{Motion Primitive $\phi^{(\text{L})}$}
\label{alg:pass_left}
\begin{lstlisting}[frame=none, framesep=2mm, mathescape=true, escapeinside=||, breaklines]
def criteria_left ($\tau$, $O$):
    time_at_each_waypoint = |\hl{extract\_time\_at\_each\_waypoint}|($\tau$)
    for t in time_at_each_waypoint:
        region = |\hl{extract\_region\_left\_of\_obs}|($O$, t)
        robot_position = |\hl{extract\_position}|($\tau$, t)
        if region|\hl{.contains}|(robot_position):
            return 1
    return 0
\end{lstlisting}
\end{algorithm}

\subsection{Pass a person from the right}
\begin{algorithm}[H]
\caption{Motion Primitive $\phi^{(\text{R})}$}
\label{alg:pass_right}
\begin{lstlisting}[frame=none, framesep=2mm, mathescape=true, escapeinside=||, breaklines]
def criteria_right ($\tau$, $O$):
    time_at_each_waypoint = |\hl{extract\_time\_at\_each\_waypoint}|($\tau$)
    for t in time_at_each_waypoint:
        region = |\hl{extract\_region\_right\_of\_obs}|($O$, t)
        robot_position = |\hl{extract\_position}|($\tau$, t)
        if region|\hl{.contains}|(robot_position):
            return 1
    return 0
\end{lstlisting}
\end{algorithm}

\subsection{Follow behind a person}
For this primitive, we consider a period of time to evaluate whether the robot has successfully followed the human, which is defined as the final few seconds before the robot reaches the goal.
\begin{algorithm}[H]
\caption{Motion Primitive $\phi^{(\text{F})}$}
\label{alg:follow}
\begin{lstlisting}[frame=none, framesep=2mm, mathescape=true, escapeinside=||, breaklines]
def criteria_follow ($\tau$, $O$):
    time_at_each_waypoint = |\hl{extract\_time\_at\_each\_waypoint}|($\tau$)
    time_period = |\hl{get\_relevant\_time\_period}|(time_at_each_waypoint)
    for t in time_period:
        region = |\hl{extract\_region\_behind\_obs}|($O$, t)
        robot_position = |\hl{extract\_position}|($\tau$, t)
        if not region|\hl{.contains}|(robot_position):
            return 0
    return 1
\end{lstlisting}
\end{algorithm}

\subsection{Yield to a person}
\begin{algorithm}[H]
\caption{Motion Primitive $\phi^{(\text{Y})}$}
\label{alg:yield}
\begin{lstlisting}[frame=none, framesep=2mm, mathescape=true, escapeinside=||, breaklines]
def criteria_yield ($\tau$, $O$):
    time_at_each_waypoint = |\hl{extract\_time\_at\_each\_waypoint}|($\tau$)
    for t in time_at_each_waypoint:
        region = |\hl{extract\_region\_in\_front\_of\_obs}|($O$, t)
        robot_position = |\hl{extract\_position}|($\tau$, t)
        if region|\hl{.contains}|(robot_position):
            return 0
    return 1
\end{lstlisting}
\end{algorithm}

\subsection{Walk through a region}
\begin{algorithm}[H]
\caption{Motion Primitive $\phi^{(\text{W})}$}
\label{alg:prefer}
\begin{lstlisting}[frame=none, framesep=2mm, mathescape=true, escapeinside=||, breaklines]
def criteria_prefer ($\tau$, $O$):
    time_at_each_waypoint = |\hl{extract\_time\_at\_each\_waypoint}|($\tau$)
    for t in time_at_each_waypoint:
        region = |\hl{extract\_region}|($O$, t)
        robot_position = |\hl{extract\_position}|($\tau$, t)
        if region|\hl{.contains}|(robot_position):
            return 1
    return 0
\end{lstlisting}
\end{algorithm}

\subsection{Avoid walking through a region}
\begin{algorithm}[H]
\caption{Motion Primitive $\phi^{(\text{A})}$}
\label{alg:avoid}
\begin{lstlisting}[frame=none, framesep=2mm, mathescape=true, escapeinside=||, breaklines]
def criteria_avoid ($\tau$, $O$):
    time_at_each_waypoint = |\hl{extract\_time\_at\_each\_waypoint}|($\tau$)
    for t in time_at_each_waypoint:
        region = |\hl{extract\_region}|($O$, t)
        robot_position = |\hl{extract\_position}|($\tau$, t)
        if region|\hl{.contains}|(robot_position):
            return 0
    return 1
\end{lstlisting}
\end{algorithm}

\section{Training Diffusion Model}
\label{sec:training_detail}

\subsection{Model Design}
We build upon a publicly available diffusion model implementation\footnote{\url{https://github.com/lucidrains/denoising-diffusion-pytorch}}. Our denoising network, $f_\theta$, consists of a 1D UNet architecture augmented with a context encoder. Each observation---whether a dynamic human or a specific region---along with the goal, is first encoded using a multilayer perceptron (MLP).
The dynamic human is represented as a predicted future trajectory, estimated under a constant velocity assumption. In contrast, a region is represented as a rectangle defined by the positions of its four corners. These embeddings are then passed through a vision transformer to generate context features, following the approach introduced in prior work~\cite{pbdiff}.


\subsection{Training}
To train the base model, we collected trajectories across three types of environments: (1) collision-free trajectories in dynamic settings, (2) trajectories that avoid specified regions, and (3) trajectories that intentionally traverse specific regions. The first two types were generated in simulation by randomly placing obstacles and planning trajectories to avoid them. For the third type, we sampled trajectories from the first two types, randomly selected a region that each trajectory passes through, and re-labeled this region as the observation to form training pairs.

We collected approximately 2 million collision-free trajectories and trained the denoising network for 2000 epochs, using a learning rate of $2 \times 10^{-4}$ and a dropout rate of 0.1. Training followed the classifier-free guidance approach~\cite{ho2021classifierfree}, where the model was conditioned on a null context (represented by zero vectors) with a probability of 20\%, instead of using features extracted from the context encoder. We also applied an exponential moving average (EMA) to the model parameters during training, which stabilizes optimization and improves generalization by smoothing out noisy updates. Following prior work~\cite{MPD}, the diffusion model performs a total of 25 denoising steps using an exponential noise schedule, generating a trajectory composed of a sequence of fixed-length, time-dependent waypoints.

To fine-tune the base denoising network for each motion primitive, we adapted the DDPO implementation\footnote{\url{https://github.com/kvablack/ddpo-pytorch}}. For simplicity, we replaced the original vision-language model (VLM)-based reward function with a heuristic-based one, designed according to the specific definitions of each motion primitive in this work. During each training epoch, we generated 32 different environments and trained the model for a total of 1000 epochs. A noteworthy observation from our experiments is that a significantly lower learning rate greatly enhances training performance. Based on this finding, we adopted a learning rate of $1 \times 10^{-6}$, which is also consistent with results reported in the literature~\cite{flare}.

\section{Additional Simulation Experiment Details}
\subsection{Simulation \ourmethod{} Setup}

We evaluate \ourmethod{} in a $20 \times 20\,\mathrm{m}^2$ 2D simulation arena. Dynamic humans are modeled as spheres, whose future positions are predicted under a constant-velocity assumption, while the static environment is represented as a rectangular region specified by the coordinates of its four corner points.
For each instruction, 20 environments are randomly initialized, assigning initial positions and speeds to the entities based on the specific requirements of the instruction. The simulation operates at a control frequency of $\Delta t = 0.1\,\mathrm{s}$, and each episode lasts for a maximum of 300 timesteps, equivalent to 30.0 seconds.

\subsection{Baseline Setup}
\label{sec:baseline_details}
In this work, we consider three baseline methods: VLM-Social-Nav~\cite{social_vlm_nav}, CoNVOI~\cite{convoi}, and BehAV~\cite{behav}. These baselines fall into two categories: the first two treat the VLM as a black-box policy that proposes a target action (e.g., next waypoint or velocity) for a geometric planner to track, while the third computes composable cost maps for planning. 

None of these baseline methods is explicitly designed to solve the problem considered in this work. Therefore, we adapt them for our experimental setup. For VLM-Social-Nav and CoNVOI, we use the latest GPT-4.1 model and disregard the high inference latency associated with invoking a remote VLM and focus on evaluating their prediction accuracy. Additionally, we modify these approaches by providing annotated screenshots of the simulates scenes and prompting the VLMs for reasoning.

For BehAV, whose core idea is to create composable costmaps and plan trajectories over them (similar to Voxposer~\cite{huang2023voxposer}), we simplify the setup by abstracting away the segmentation vision model. Instead, we provide BehAV with ground-truth annotations obtained directly from the simulation environment and evaluate solely how well costmap-based planning enables instruction following.

For each baseline method, we conducted a grid search over various combinations of motion planner hyperparameters—specifically, different weights of the constituent cost functions—using a small tuning set. We then selected the hyperparameters that achieved the highest overall success rates. 

\subsection{Supplementary Quantitative Results}
\label{sec:pretrain_ft_quantitative}

\subsubsection{Learning Motion Primitives}

\begin{wraptable}{r}{0.6\textwidth}  
\scriptsize
\centering
\caption{Comparison of Fine-tuned and Pre-trained Diffusion Models for Representing Motion Primitives.}
\label{tab:pretrain_finetune}
\begin{tabular}{@{}c@{\hspace{2pt}}
c@{\hspace{3pt}}
c@{\hspace{3pt}}
c@{\hspace{3pt}}
c@{\hspace{5pt}}
c@{\hspace{3pt}}
c@{\hspace{3pt}}
c@{\hspace{3pt}}
c@{}}
\toprule
\textbf{MP} & 
\multicolumn{4}{c}{\textbf{Pre-trained Model}} & 
\multicolumn{4}{c}{\textbf{Fine-tuned Model}} \\ 
\cmidrule(r){2-5} \cmidrule(r){6-9}
& 
\textbf{SR}(\%)$\uparrow$ & \textbf{IA}(\%)$\uparrow$ & \textbf{CF}(\%)$\uparrow$ & \textbf{GR}(\%)$\uparrow$ & 
\textbf{SR}(\%)$\uparrow$ & \textbf{IA}(\%)$\uparrow$ & \textbf{CF}(\%)$\uparrow$ & \textbf{GR}(\%)$\uparrow$ \\ 
\midrule
L & 44.0 & 44.0 & 100.0 & 100.0 & 100.0 & 100.0 & 100.0 & 100.0 \\
R & 37.0 & 37.0 & 100.0 & 100.0 & 100.0 & 100.0 & 100.0 & 100.0 \\
F & 27.0 & 27.0 & 100.0 & 100.0 & 99.0 & 100.0 & 100.0 & 99.0 \\
Y & 48.0 & 48.0 & 100.0 & 100.0 & 100.0 & 100.0 & 100.0 & 100.0 \\
W & 34.0 & 34.0 & 100.0 & 100.0 & 100.0 & 100.0 & 100.0 & 100.0 \\
A & 100.0 & 100.0 & 100.0 & 100.0 & 100.0 & 100.0 & 100.0 & 100.0 \\
\bottomrule
\end{tabular}
\end{wraptable}


Beyond demonstrating that the RL fine-tuning procedure enables the learning of effective motion primitives, we further analyze results obtained from comparing the pre-trained and fine-tuned models. As shown in \tabref{pretrain_finetune}, the pre-trained model, which is trained to generate collision-free, goal-reaching trajectories, consistently achieves these objectives across all evaluated motion primitives. In particular, the pre-trained model achieves a 100\% success rate for the \emph{``Avoid walking through a region''} motion primitive. This outcome is anticipated, as avoiding designated regions is closely aligned with the collision avoidance objective emphasized during the pre-training phase.

\begin{table}[t]
\scriptsize
\caption{Detailed Simulation Evaluation Results}
\label{tab:simulation_result}
\centering
\begin{tabularx}{\linewidth}{
@{}c@{}c
@{\hspace{7pt}}c@{\hspace{7pt}}c@{\hspace{7pt}}c@{\hspace{7pt}}c
@{\hspace{7pt}}c@{\hspace{7pt}}c@{\hspace{7pt}}c@{\hspace{7pt}}c
@{\hspace{7pt}}c@{\hspace{7pt}}c@{\hspace{7pt}}c@{\hspace{7pt}}c
@{\hspace{7pt}}c@{\hspace{7pt}}c@{\hspace{7pt}}c@{\hspace{7pt}}c
}
\toprule
\multirow{3}{*}{\# }& \multirow{3}{*}{Combination}& 
\multicolumn{8}{c}{VLM-based Policy} & 
\multicolumn{4}{c}{Compose Costmaps} & 
\multicolumn{4}{c}{Compose Primitives} \\
\cmidrule(lr){3-10} \cmidrule(lr){11-14} \cmidrule(lr){15-18}
&  & 
\multicolumn{4}{c}{VLM-Social-Nav (\%)} & 
\multicolumn{4}{c}{Convoi (\%)} & 
\multicolumn{4}{c}{Behav (\%)} & 
\multicolumn{4}{c}{ComposableNav (ours) (\%)} \\
\cmidrule(lr){3-6} \cmidrule(lr){7-10} \cmidrule(lr){11-14} \cmidrule(lr){15-18}
&  & 
\textbf{SR}$\uparrow$ & \textbf{IA}$\uparrow$ & \textbf{CF}$\uparrow$ & \textbf{GR}$\uparrow$ & 
\textbf{SR}$\uparrow$ & \textbf{IA}$\uparrow$ & \textbf{CF}$\uparrow$ & \textbf{GR}$\uparrow$ & 
\textbf{SR}$\uparrow$ & \textbf{IA}$\uparrow$ & \textbf{CF}$\uparrow$ & \textbf{GR}$\uparrow$ & 
\textbf{SR}$\uparrow$ & \textbf{IA}$\uparrow$ & \textbf{CF}$\uparrow$ & \textbf{GR}$\uparrow$ \\

\midrule
\multirow{6}{*}{\rotatebox{90}{\textbf{1 Primitive}}} 
& L & 
55.0 & 55.0 & 100.0 & 100.0 & 
70.0 & 75.0 & 95.0 & 100.0 & 
65.0 & 65.0 & 100.0 & 100.0 & 
\textbf{100.0} &  100.0 & 100.0 & 100.0  \\
& R & 
55.0 & 55.0 & 100.0 & 100.0 & 
70.0 & 75.0 & 95.0 & 100.0 & 
65.0 & 65.0 & 100.0 & 100.0 & 
\textbf{100.0} &  100.0 & 100.0 & 100.0  \\
& F & 
5.0 & 5.0 & 100.0 & 100.0 & 
0.0 & 0.0 & 95.0 & 100.0 & 
70.0 &  75.0 & 100.0 & 95.0 & 
\textbf{99.0} & 100.0 & 100.0 & 99.0   \\
& Y & 
25.0 & 25.0 & 95.0 & 100.0 & 
40.0 & 40.0 & 95.0 & 100.0 & 
100.0 & 100.0 & 100.0 & 100.0 & 
\textbf{100.0} &  100.0 & 100.0 & 100.0  \\
& W & 
0.0 & 0.0 & 100.0 & 100.0 & 
55.0 & 55.0 & 100.0 & 100.0 & 
55.0 & 55.0 & 100.0 & 100.0 & 
\textbf{100.0} &  100.0 & 100.0 & 100.0  \\
& A & 
0.0 & 0.0 & 100.0 & 100.0 & 
100.0 & 100.0 & 100.0 & 100.0 & 
100.0 & 100.0 & 100.0 & 100.0 & 
\textbf{100.0} &  100.0 & 100.0 & 100.0  \\
\rowcolor{blue!10}
\multicolumn{1}{>{\columncolor{white}}c}{} & Overall & 23.3 & 23.3 & 99.2 & 100.0 & 55.8 & 57.5 & 96.7 & 100.0 & 75.8 & 76.7 & 100.0 & 99.2 & \textbf{99.8}  & 100.0 & 100.0 & 99.8\\
\midrule
\multirow{8}{*}{\rotatebox{90}{\textbf{2 Motion Primitives}}} 
& L+R & 
5.0 & 5.0 & 100.0 & 100.0 & 
0.0 & 0.0 & 100.0 & 100.0 & 
0.0 & 0.0 & 100.0 & 100.0 & 
\textbf{33.0} &  34.0 & 86.0 & 100.0  \\
& P+F & 
10.0 & 10.0 & 100.0 & 100.0 & 
5.0 & 5.0 & 90.0 & 100.0 & 
0.0 & 0.0 & 90.0 & 100.0 & 
\textbf{85.0} & 85.0 & 99.0 & 100.0   \\
& Y+P & 
20.0 & 20.0 & 100.0 & 100.0 & 
55.0 & 55.0 & 95.0 & 100.0 & 
90.0 &  90.0 & 100.0 & 100.0 & 
\textbf{90.0} & 93.0 & 97.0 & 100.0   \\
& Y+F & 
5.0 & 5.0 & 95.0 & 100.0 & 
0.0 & 0.0 & 80.0 & 95.0 & 
85.0 & 85.0 & 100.0 & 100.0 & 
\textbf{85.0} & 85.0 & 100.0 & 100.0  \\
& W+P & 
0.0 & 0.0 & 100.0 & 100.0 & 
20.0 & 20.0 & 95.0 & 100.0 & 
5.0 & 5.0 & 100.0 & 100.0 & 
\textbf{60.0} &  61.0 & 97.0 & 100.0  \\
& W+Y & 
0.0 & 0.0 & 100.0 & 100.0 & 
35.0 & 35.0 & 95.0 & 100.0 & 
40.0 & 40.0 & 100.0 & 100.0 & 
\textbf{99.0}  & 99.0 & 100.0 & 100.0   \\
& A+P & 
0.0 & 0.0 & 90.0 & 100.0 & 
65.0 & 65.0 & 100.0 & 100.0 & 
50.0 & 50.0 & 100.0 & 95.0 & 
\textbf{67.0} & 68.0 & 99.0 & 100.0   \\
& A+F & 
0.0 & 0.0 & 100.0 & 100.0 & 
0.0 & 0.0 & 100.0 & 100.0 & 
35.0 & 40.0 & 100.0 & 35.0 & 
\textbf{85.0} &  85.0 & 100.0 & 100.0  \\
\rowcolor{blue!10}
\multicolumn{1}{>{\columncolor{white}}c}{} & Overall & 5.0 & 5.0 & 98.1 & 100.0 & 22.5 &  22.5 & 94.4 & 99.4 & 38.1 & 38.8 & 98.8 & 91.2 & \textbf{75.5}  & 76.2 & 97.2 & 100.0\\
\midrule
\multirow{8}{*}{\rotatebox{90}{\textbf{3 Motion Primitives}}} 
& P+F+Y &  
0.0 & 0.0 & 100.0 & 100.0 & 
5.0 & 5.0 & 85.0 & 100.0 & 
0.0 & 0.0 & 100.0 & 100.0 & 
\textbf{30.0} &  34.0 & 91.0 & 100.0  \\
& P+F+W & 
0.0 & 0.0 & 100.0 & 100.0 & 
5.0 & 5.0 & 80.0 & 100.0 & 
0.0 & 0.0 & 100.0 & 95.0 & 
\textbf{58.0} & 61.0 & 92.0 & 100.0   \\
& P+Y+W & 
0.0 & 0.0 & 100.0 & 100.0 & 
20.0 & 20.0 & 90.0 & 100.0 & 
5.0 & 5.0 & 100.0 & 90.0 & 
\textbf{38.0} & 42.0 & 92.0 & 100.0   \\
& W+W+Y & 
0.0 & 0.0 & 100.0 & 100.0 & 
10.0 &  10.0 & 90.0 & 100.0 & 
5.0 & 5.0 & 100.0 & 50.0  & 
\textbf{87.0} & 87.0 & 100.0 & 100.0   \\
& A+A+Y & 
0.0 & 0.0 & 90.0 & 100.0 & 
5.0  & 5.0 & 100.0 & 95.0 & 
15.0 & 75.0 & 100.0 & 15.0 & 
\textbf{86.0} & 86.0 & 99.0 & 100.0   \\
& A+W+Y & 
0.0  & 0.0 & 100.0 & 100.0 & 
\textbf{20.0}  & 20.0 & 100.0 & 100.0 & 
10.0  & 10.0 & 100.0 & 60.0 & 
0.0  & 0.0 & 99.0  & 100.0  \\
& A+P+F & 
0.0  & 0.0 & 100.0 & 100.0 & 
15.0  & 15.0 & 95.0 &  100.0 & 
0.0  & 0.0 & 100.0 &  90.0 & 
\textbf{93.0}  & 95.0 & 94.0 & 100.0  \\
& A+W+F & 
0.0  & 0.0 & 100.0 & 40.0 & 
5.0  & 5.0 & 95.0 &  40.0 & 
45.0 & 50.0 & 90.0 & 85.0 & 
\textbf{77.0} & 77.0 & 98.0 & 100.0\\
\rowcolor{blue!10}
\multicolumn{1}{>{\columncolor{white}}c}{} & Overall & 0.0 & 0.6 & 98.8 & 92.5 & 10.6 & 10.6 & 91.9 & 91.9 & 10.0 & 18.1 & 98.8& 80.6 & \textbf{58.6} & 60.2 & 95.6 & 100.0 \\
\midrule
\multirow{8}{*}{\rotatebox{90}{\textbf{4 Motion Primitives}}} 
& A+W+F+Y & 
0.0 & 0.0 & 90.0 & 60.0 & 
0.0. & 0.0 & 90.0 & 60.0 & 
30.0 & 30.0 & 95.0 & 60.0 & 
\textbf{33.0} &  35.0 & 94.0 & 98.0  \\
& A+W+F+P & 
0.0 & 5.0 & 90.0 & 50.0 & 
0.0 & 0.0 & 85.0 & 50.0 & 
0.0 & 0.0 & 95.0 & 10.0 & 
\textbf{59.0} & 61.0 & 72.0 & 100.0   \\
& A+W+A+Y & 
0.0. & 0.0 & 100.0 & 50.0 & 
35.0. & 35.0 & 100.0 & 70.0 & 
2.0. & 20.0 & 100.0 & 25.0 & 
\textbf{46.0} & 46.0 & 76.0 & 100.0 \\
& W+W+Y+A & 
0.0. & 0.0 & 90.0 & 100.0 & 
5.0. & 5.0 & 65.0 & 100.0 & 
0.0. & 0.0 & 85.0 & 100.0 & 
\textbf{71.0} & 78.0 & 86.0 & 100.0 \\
& W+P+Y+A & 
0.0. & 0.0 & 70.0 & 100.0 & 
25.0. & 25.0 & 75.0 & 100.0 & 
5.0. & 5.0 & 70.0 & 100.0 & 
\textbf{28.0} & 34.0 & 78.0 & 100.0 \\
& W+P+F+A & 
0.0. & 0.0 & 80.0 & 80.0 & 
0.0. & 0.0 & 75.0 & 100.0 & 
0.0. & 0.0 & 80.0 & 100.0 & 
\textbf{24.0} & 31.0 & 81.0 & 100.0 \\
& P+F+Y+A & 
0.0. & 0.0 & 80.0 & 100.0 & 
0.0. & 0.0 & 85.0 & 100.0 & 
0.0. & 0.0 & 100.0 & 100.0 & 
\textbf{18.0} & 23.0 & 82.0 & 95.0 \\
& A+W+Y+A & 
0.0. & 0.0 & 60.0 & 100.0 & 
0.0. & 0.0 & 75.0 & 100.0 & 
0.0. & 0.0 & 90.0 & 55.0 & 
0.0. & 0.0 & 92.0 & 99.0 \\
\rowcolor{blue!10}
\multicolumn{1}{>{\columncolor{white}}c}{} & Overall & 0.0 & 0.6 & 82.5 & 80.0 & 8.1 & 8.1 & 81.2 & 84.4 & 6.9 & 6.9 & 89.4 & 68.8 & \textbf{34.9} & 38.5 & 82.6 & 99.0  \\
\bottomrule
\end{tabularx}
\vspace{-2em}
\end{table}
\subsubsection{Composing Motion Primitives}
\label{sec:full_simulation_quantitative}

\begin{figure}[H]
    \centering
    \includegraphics[width=1\linewidth]{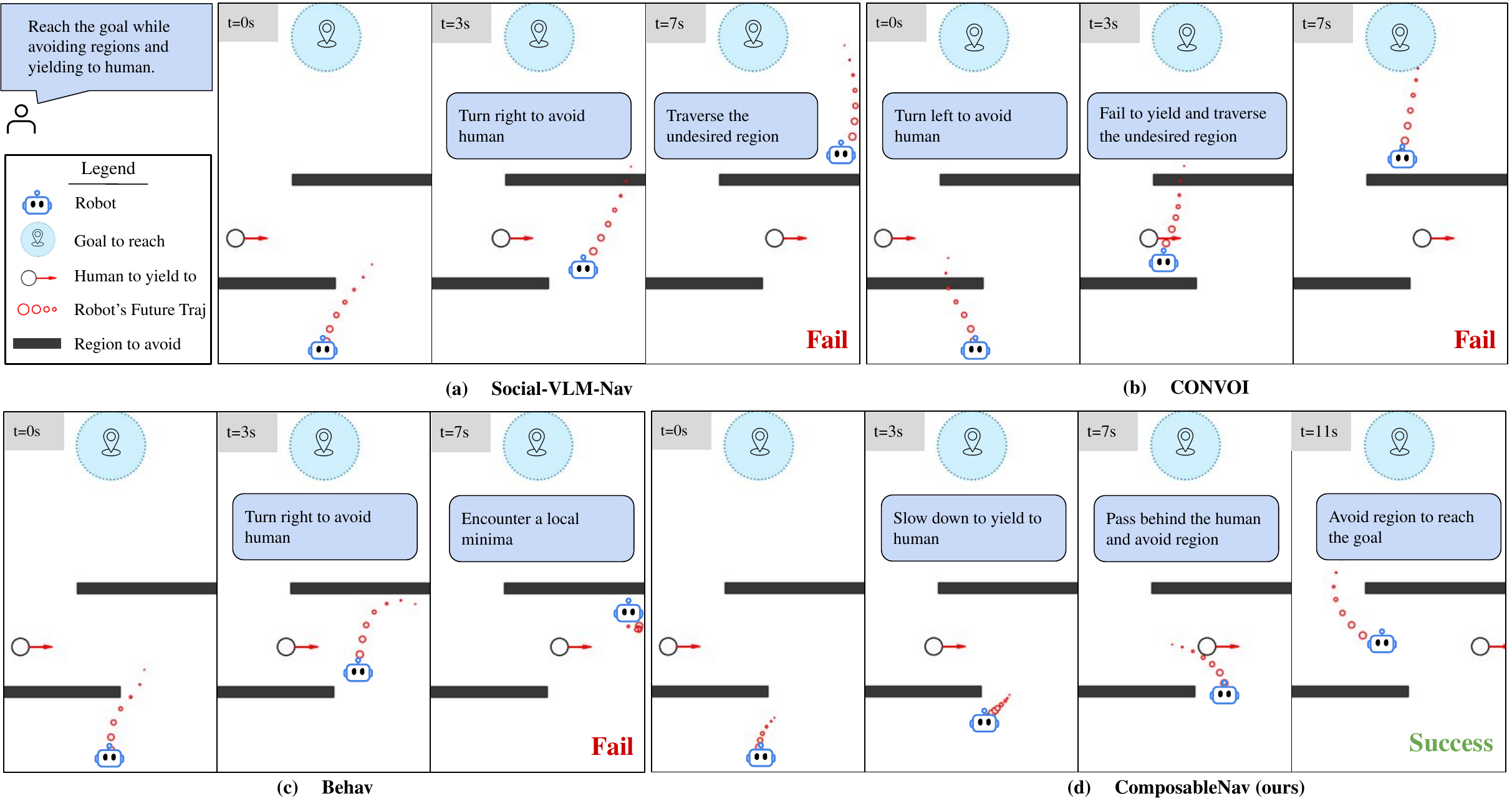}
    \caption{Qualitative Simulation Results}
    \label{fig:simulation_illustration}
\end{figure}

We present the quantitative results of 24 motion primitive combinations in our simulation testbed, as detailed in \tabref{simulation_result}. Each combination (e.g., ``L+R") corresponds to a specific natural language instruction (e.g., ``Pass person 1 from the left and person 2 from the right"), which maps to a distinct robot motion trajectory illustrated in \figref{first_figure}. Given the similarity between the primitives ``pass a person from the left" and ``pass a person from the right"—which differ only in direction—we unify them under a general instruction category: ``Pass a person" (denoted as P) for motion composition purposes. Accordingly, we evaluate our method, \ourmethod{}, on its ability to handle both left and right variants using a single instruction specification.

The quantitative results are summarized in \tabref{simulation_result}. Across all testbed scenarios, \ourmethod{} consistently outperforms all baseline methods in terms of success rate, particularly as the number of motion primitives in an instruction increases. While baseline methods perform reasonably with simple instructions, their performance deteriorates notably with more complex instruction sets.

Methods relying on Vision-Language Models (VLMs) as black-box policies perform particularly poorly. These models are not designed for such navigation tasks and often fail to maintain planning consistency, especially as instruction complexity grows. Similarly, BehAV performs adequately with one or two motion primitives but suffers as composition complexity increases. In addition, BehAV has the lowest goal-reaching rate, which suggests that such a costmap-based method tends to get trapped in local minima and is unable to complete tasks within the allotted time.

Furthermore, baseline methods exhibit high variance in performance across different instruction combinations. In many cases, they fail to generate any viable, instruction-following trajectory. In contrast, \ourmethod{}—\emph{despite not being explicitly trained for any possible instruction composition}—demonstrates strong generalization capabilities and consistently higher success rates across a wide range of scenarios.

To complement the quantitative analysis, we provide a qualitative illustration in \figref{simulation_illustration}. Here, we observe that while VLM-based methods may initially steer the robot in the correct direction, they lack the responsiveness and consistency needed for sustained instruction following. These methods often begin to avoid regions or yield to pedestrians, but then fail to complete subsequent specifications. The BehAV method often gets stuck in local minima and fails to reach the goal within the allowed time. In contrast, \ourmethod{} effectively produces instruction-aligned behaviors, simultaneously satisfying all given specifications—such as avoiding restricted regions and yielding to oncoming pedestrians.

\section{Real-World Deployment}

\subsection{Robot Setup}
\label{sec:robot_setup}
\label{sec:latency}

\begin{wrapfigure}{r}{0.4\textwidth}
    \centering
    \includegraphics[width=0.3\textwidth]{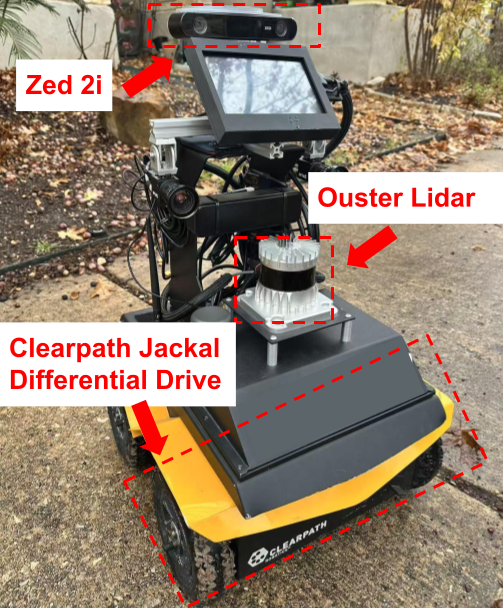}
    \caption{Robot Setup.}
    \label{fig:robot_setup}
\end{wrapfigure}
We deploy \ourmethod{} on a Clearpath Jackal robot equipped with a Zed 2i camera for human tracking and an Ouster LiDAR for generating point cloud data to create an obstacle map for collision avoidance, as shown in \figref{robot_setup}. For localization, we apply ENML~\cite{enml}, which provides robust position estimation. 
The system is built using ROS and consists of a navigation stack with three main modules: a perception module, a diffusion planning module, and an MPC motion planning module. 
The perception module leverages Zed's internal human-tracking algorithm to detect and track humans while using Ouster’s point cloud data to detect obstacles for collision avoidance. 
The diffusion planning module loads diffusion models corresponding to various motion primitives and composes the appropriate models based on instructions and environmental observations. The MPC motion planning module tracks the time-dependent trajectory generated by the diffusion planner and computes real-time motion control commands. All computations are executed entirely onboard, utilizing an Intel i7-9700TE CPU and an NVIDIA RTX A2000 GPU.

\subsection{Real-time Deployment}
\label{sec:real_time_deploy}

In our deployment, the motion controller and trajectory replanning run at fixed frequencies: the motion controller operates at $10\,\text{Hz}$ (every $0.1\,\text{s}$), while trajectory replanning executes at $0.67\,\text{Hz}$ (every $1.5\,\text{s}$). We define \textit{real-time} operation as producing outputs within the motion control cycle of $0.1\,\text{s}$, ensuring both control and replanning complete within this bound. Below, we describe the implementation details that enable \ourmethod{} to meet this requirement with onboard hardware.

\subsubsection{Real-time Motion Control}
To produce motion commands in real-time, we employ a Model Predictive Path Integral (MPPI)~\cite{mppi}, a sampling-based MPC controller, to track the time-dependent trajectories generated by the composed diffusion models. During navigation, MPPI uses a differential drive kinematic model to predict the robot’s future states and minimizes the deviation between these predictions and the target trajectory over a short planning horizon. It also enforces constraints on acceleration and velocity to guarantee feasible and safe control inputs.

\subsubsection{Real-time Trajectory Replanning}
To enable trajectory replanning on only the robot’s onboard compute, we adopt the adaptive online replanning framework introduced in prior work~\cite{adaptive_online_replanning}, specifically the \textit{Replan from Previous Context} method. The core insight is that the current trajectory is typically close to optimal, so instead of discarding it and replanning from scratch, \ourmethod{} perturbs the trajectory by applying a few forward diffusion steps $q(x_t \mid x_{t-1})$ and then partially denoises it to generate an updated trajectory conditioned on the latest observations.

In practice, we apply five forward diffusion steps to inject noise into the current trajectory, followed by five reverse diffusion steps to denoise it. During this process, the states already visited by the robot are fixed, and only the future segments are updated.

We introduce two key optimizations to ensure efficiency:
(1) Since all fine-tuned diffusion models share an identical architecture derived from a common base, we leverage PyTorch’s vectorized mapping operation (\texttt{vmap}) to batch their execution in parallel.  
(2) We further accelerate inference using PyTorch’s compilation feature (\texttt{torch.compile}).  

With these optimizations, \ourmethod{} achieves real-time replanning entirely on onboard compute, with latency results summarized in \tabref{latency_results}.

\begin{table}[H]
    \centering
    \caption{\ourmethod{} Inference Latency on Robot Hardware}
    \label{tab:latency_results}
    \begin{tabular}{@{}l@{\hspace{4pt}}c@{\hspace{6pt}}c@{\hspace{6pt}}c@{\hspace{6pt}}c@{}}
    \toprule
    \multirow{2}{*}{\textbf{Latency}} & \multicolumn{4}{c}{\textbf{\# of Composed MPs}} \\ 
     \cmidrule(r){2-5}
    & \textbf{1} & \textbf{2} & \textbf{3} & \textbf{4}\\
    \midrule
    \textbf{Initial Plan (s)$\downarrow$} & 0.144 $\pm$ 0.014 & 0.243 $\pm$ 0.009 & 0.329 $\pm$ 0.014 & 0.413 $\pm$ 0.010\\
    \textbf{Replan (s)$\downarrow$} & 0.027 $\pm$ 0.002 & 0.036 $\pm$ 0.004 & 0.049 $\pm$ 0.003 & 0.060 $\pm$ 0.005\\
    \bottomrule
    \end{tabular}
\end{table}

\section{Additional Robot Deployment Details}
\label{sec:additional_robot_deployment}

\subsection{Quantitative Experiment Details}
\begin{figure}[H]
    \centering
    \includegraphics[width=\linewidth]{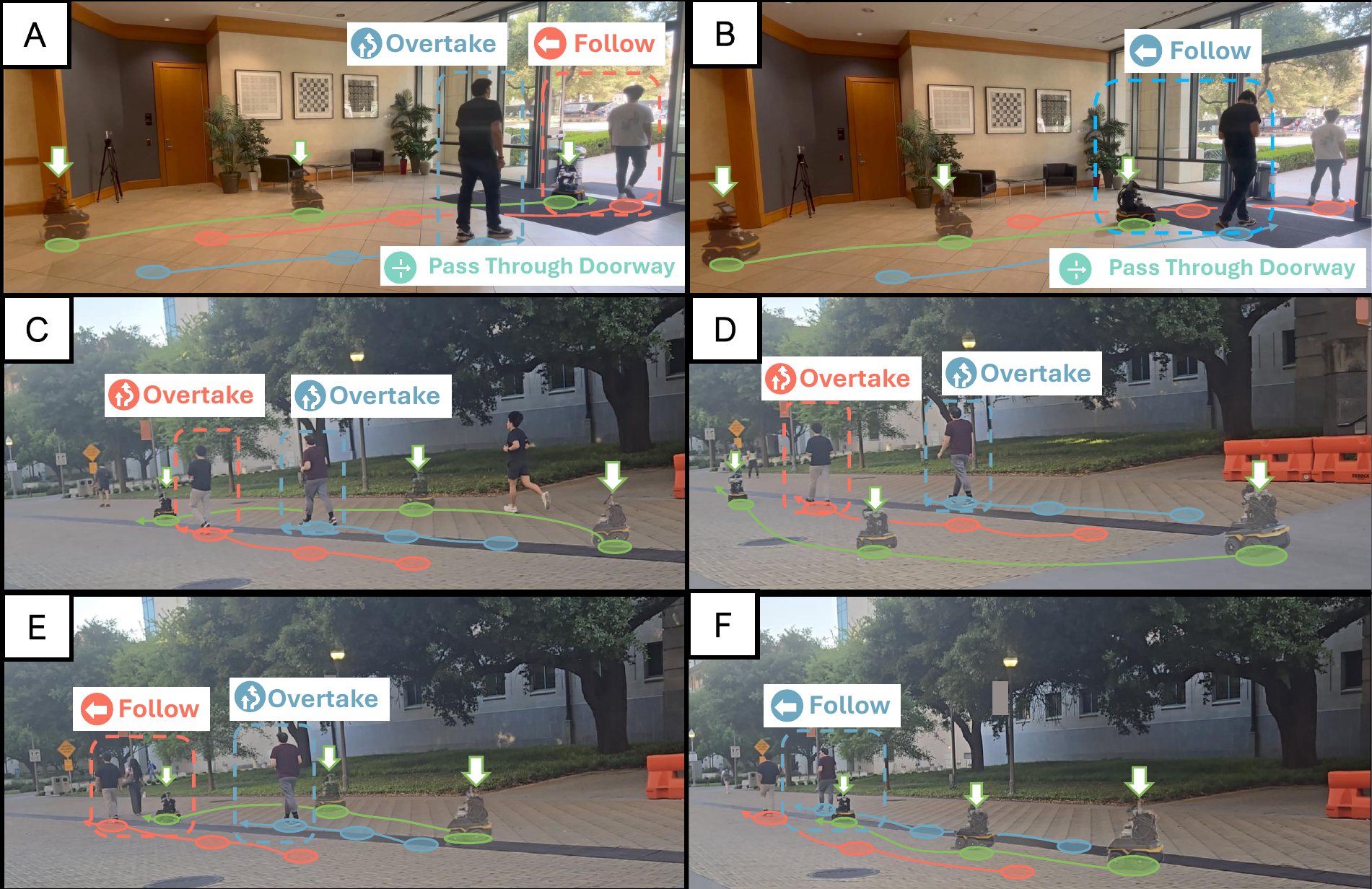}
    \caption{Quantitative Experiment Illustration}
    \label{fig:real-world-exp}
\end{figure}
We evaluate \ourmethod{} in two common real-world scenarios: navigating through a narrow doorway and walking outdoors in an open environment, using a total of six instructions—two for the doorway scenario and four for the outdoor scenario, as illustrated in \figref{real-world-exp}.

\paragraph{Doorway Instructions:}
\begin{itemize}
    \item \textbf{Instruction 1:} ``Go through the doorway after the person wearing a black shirt and before the person wearing a white shirt'' (\figref{real-world-exp}A)
    \item \textbf{Instruction 2:} ``Go through the doorway before the person wearing a black shirt'' (\figref{real-world-exp}B)
\end{itemize}

\paragraph{Outdoor Instructions:}
\begin{itemize}
    \item \textbf{Instruction 3:} ``Pass both the person wearing a black shirt and a maroon shirt from the right side of the road'' (\figref{real-world-exp}C)
    \item \textbf{Instruction 4:} ``Pass both the person wearing a black shirt and a maroon shirt from the left side of the road'' (\figref{real-world-exp}D)
    \item \textbf{Instruction 5:} ``Pass the person wearing a maroon shirt and follow the person wearing a black shirt'' (\figref{real-world-exp}E)
    \item \textbf{Instruction 6:} ``Follow the person wearing a black shirt'' (\figref{real-world-exp}F)
\end{itemize}

Each instruction was tested over 10 trials, and we report the corresponding success rates in \figref{real-world-exp}. These experiments demonstrate how \ourmethod{} enables customizable robot behavior that aligns with human preferences. For instance, in the doorway scenario, a human operator might prefer the robot to be polite by following the person in a black shirt, or alternatively, instruct it to hurry and enter after the person in a white shirt. In the outdoor scenario, preferences may include keeping to a specific side of the road to follow the human flow, or adjusting the robot's pace to follow a specific individual, such as someone in a maroon or black shirt.

\subsection{Additional Real-World Deployment Demo}
\begin{figure}[H]
    \centering
    \includegraphics[width=0.9\linewidth]{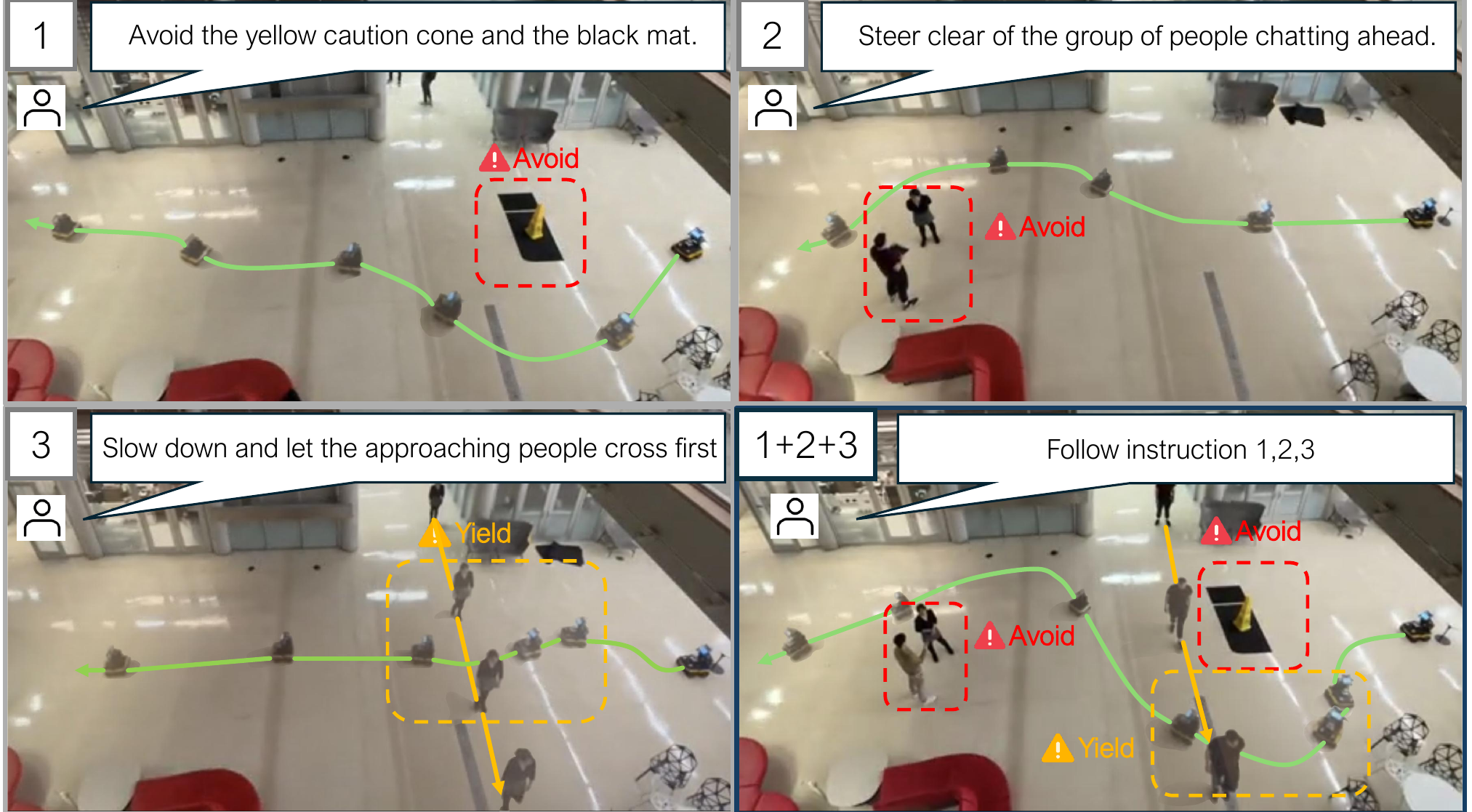}
    \caption{Real-World Composition Experiment}
    \label{fig:real_world_compose}
\end{figure}

\begin{figure}[H]
    \centering
    \begin{subfigure}[b]{0.45\textwidth}
        \centering
        \includegraphics[width=\linewidth]{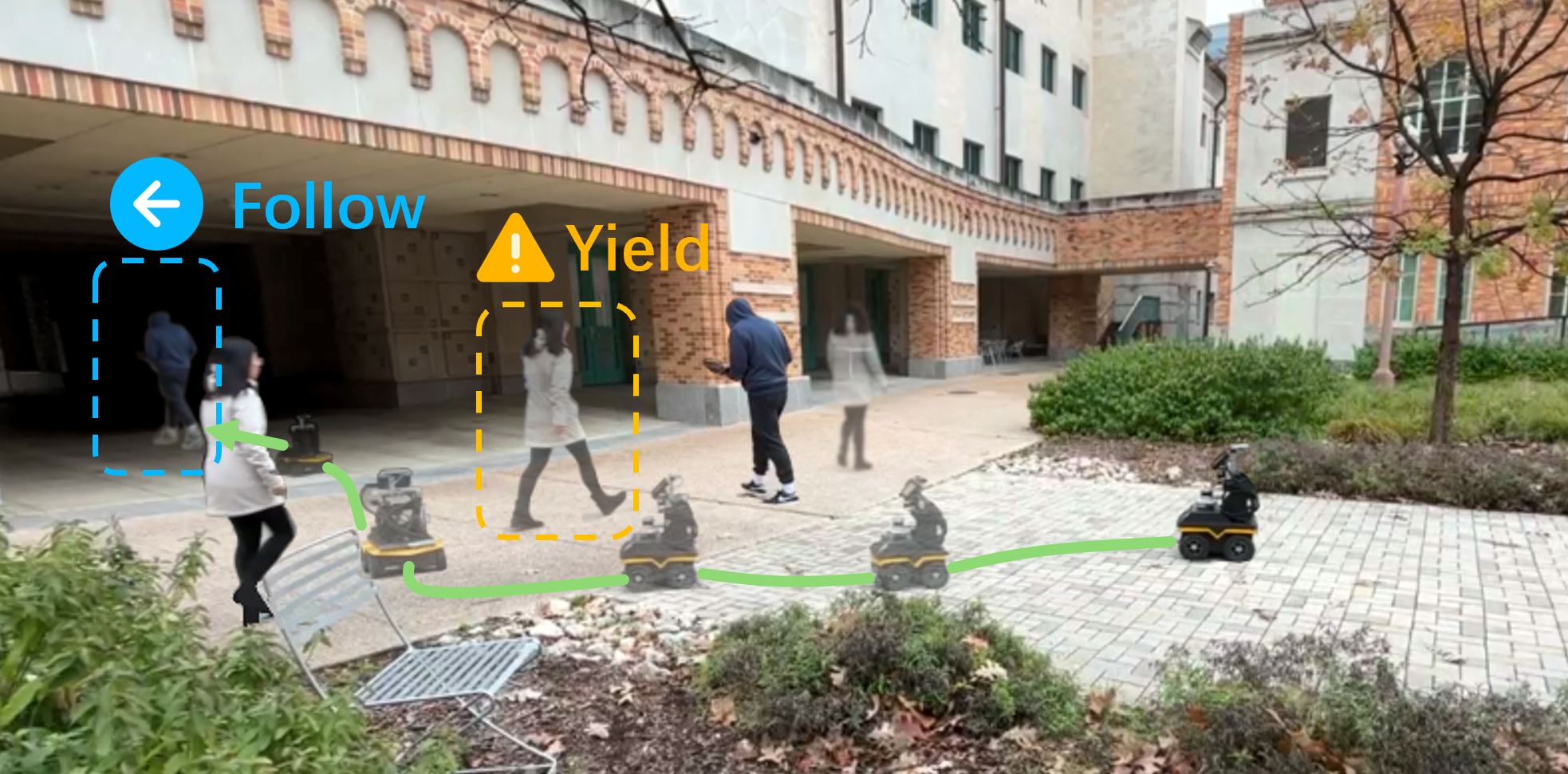}
        \caption{Yield to the person in the white coat and follow the person wearing a dark hoodie.}
        \label{fig:first}
    \end{subfigure}
    \hfill
    \begin{subfigure}[b]{0.45\textwidth}
        \centering
        \includegraphics[width=\linewidth]{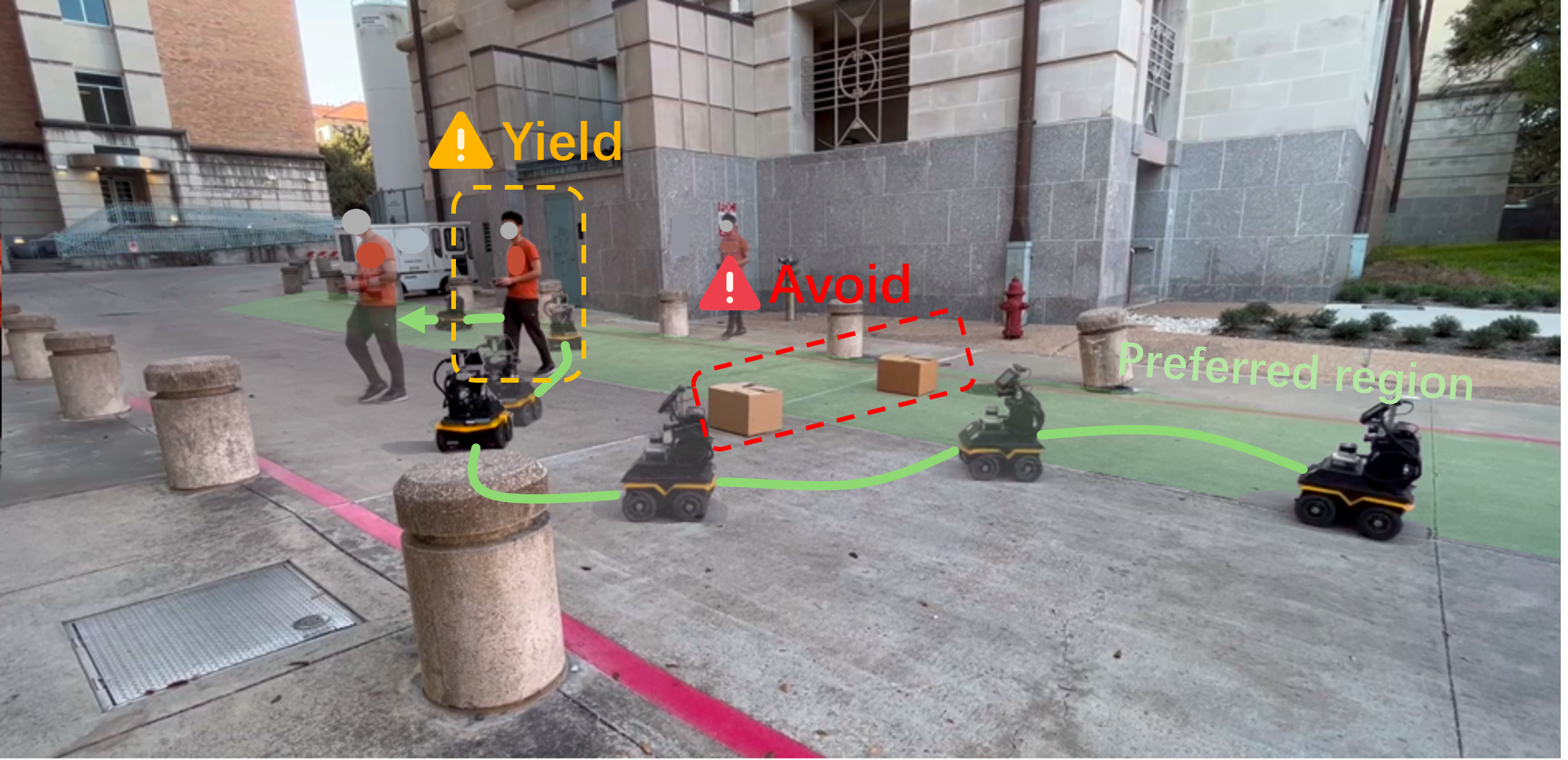}
        \caption{Avoid walking between the boxes, yield to the person, and stay on the right side of the road.}
        \label{fig:second}
    \end{subfigure}
    


\end{figure}

\subsection{Deployment Failure Case Analysis}
We conducted a qualitative analysis of the failure cases observed when deploying \ourmethod{} on the robot and identified two common issues. The first issue stems from human tracking errors. Since both the robot and the human are in continuous motion, the person may temporarily exit the camera's field of view—particularly when the robot turns—causing the system to lose track of them, even if they later reappear. While we applied a simple nearest-neighbor heuristic to reassign the human based on previous tracking data, occasional failures still occur, where the robot is unable to reliably re-identify the person. The second issue arises during replanning. We observed that the newly generated plan can sometimes diverge significantly from the original one. This can lead the MPPI controller to issue large acceleration or deceleration commands, resulting in jerky movements. Consequently, the robot may overshoot its intended state and struggle to stay on the planned path. We plan to address these issues in future work

\end{document}